%% file: submission.tex
\documentclass[11pt]{article}

\PassOptionsToPackage{hyphens}{url}   
\usepackage[preprint]{acl}

\usepackage{fontspec}          
\defaultfontfeatures{Ligatures=TeX}
\usepackage{latexsym}
\usepackage{microtype}
\usepackage{graphicx}
\usepackage{amssymb}        
\usepackage{booktabs}       
\usepackage{multirow}       
\usepackage{multicol}       
\usepackage{xcolor}         
\usepackage{colortbl}       
\usepackage{listings}       
\usepackage{todonotes}      
\usepackage{subcaption}     
\usepackage{rotating}       
\usepackage{array}          
\usepackage{footnote}       
\usepackage{url}            
\usepackage{longtable}      
\usepackage{pdflscape}      
\usepackage{tikz}           
\usetikzlibrary{arrows.meta,positioning,backgrounds}

\definecolor{codebg}{rgb}{0.965,0.970,0.990}    
\definecolor{coderule}{rgb}{0.50,0.62,0.78}     
\definecolor{codekey}{rgb}{0.08,0.20,0.64}      
\definecolor{codecomment}{rgb}{0.38,0.44,0.52}  
\definecolor{codestring}{rgb}{0.10,0.48,0.22}   
\definecolor{codebuiltin}{rgb}{0.52,0.12,0.56}  

\lstdefinestyle{pythonstyle}{
  backgroundcolor=\color{codebg},
  basicstyle=\ttfamily\footnotesize,
  keywordstyle=\color{codekey}\bfseries,
  keywordstyle=[2]\color{codebuiltin}\bfseries, 
  commentstyle=\color{codecomment}\itshape,
  stringstyle=\color{codestring},
  breaklines=true,
  breakatwhitespace=false,
  postbreak=\mbox{\textcolor{coderule}{$\hookrightarrow$}\space},
  frame=lines,               
  framesep=5pt,
  rulecolor=\color{coderule},
  language=Python,
  morekeywords=[2]{True,False,None,self,cls,print,len,range,
                   enumerate,zip,sum,list,dict,str,int,float},
  showstringspaces=false,
  keepspaces=true,
  tabsize=4,
  aboveskip=7pt,
  belowskip=3pt,
  xleftmargin=5pt,
  xrightmargin=5pt,
}
\newcommand{\tklyes}{\textcolor{green!60!black}{\checkmark}}
\newcommand{\tklno}{\textcolor{red!70!black}{$\times$}}
\newcommand{\tklpar}{\textcolor{orange!80!black}{$\sim$}}  

\title{TurkicNLP: A Script-Aware Toolkit for Multilingual and\\Cross-Lingual Processing of 20+ Turkic Languages}

\author{Sherzod Hakimov \\
  University of Potsdam \\
  \texttt{firstnamelastname@gmail.com}}

\begin{document}
\maketitle

\begin{abstract}
\input{sections/abstract}
\end{abstract}

\input{sections/01-introduction}
\input{sections/02-related-work}
\input{sections/03-design}
\input{sections/04-scripts}
\input{sections/05-processors}
\input{sections/06-evaluation}
\input{sections/10-conclusion}
\input{sections/08-future}
\input{sections/09-limitations}

\sloppy
\clearpage
\bibliography{custom,anthology_0,anthology_1}

\appendix
\input{sections/appendix-d-resources}
\input{sections/llm-models-table}
\input{sections/appendix-e-challenge-sentences}
\input{sections/appendix-f-script-results}
\input{sections/appendix-g-morphseg}

\end{document}

%% file: sections/abstract.tex
Turkic languages are spoken by about 200 million people and are written in Latin, Cyrillic, and Perso-Arabic scripts, yet the NLP tools that exist for them are scattered across languages, scripts, and incompatible software. We present TurkicNLP, an open-source Python library that offers a single pipeline for 24 Turkic languages, covering tokenization, morphological analysis, POS tagging, dependency parsing, named entity recognition, language identification, sentence embeddings, and machine translation. The library is designed for cross-lingual use: it detects and converts between scripts, maps 23 languages into the shared Common Turkic Alphabet, and includes multilingual models that share parameters across the family and transfer to languages without annotated data. Our evaluation shows the benefits of this design. The multilingual parser outperforms per-language parsers, the multilingual morphological analyzer complements finite-state analyzers where their lexicons are small, and both produce usable analyses for languages unseen in training. Transliteration is almost fully reversible, and converting text to the Common Turkic Alphabet removes most of the additional tokenization cost that large language models incur on Cyrillic- and Arabic-script Turkic languages. Code, models, and documentation are available at \url{https://github.com/turkic-nlp/turkicnlp}.

%% file: sections/01-introduction.tex
\section{Introduction}
\label{sec:intro}

\subsection{Background}
\begin{figure*}[!t]
    \centering
    \includegraphics[width=\textwidth]{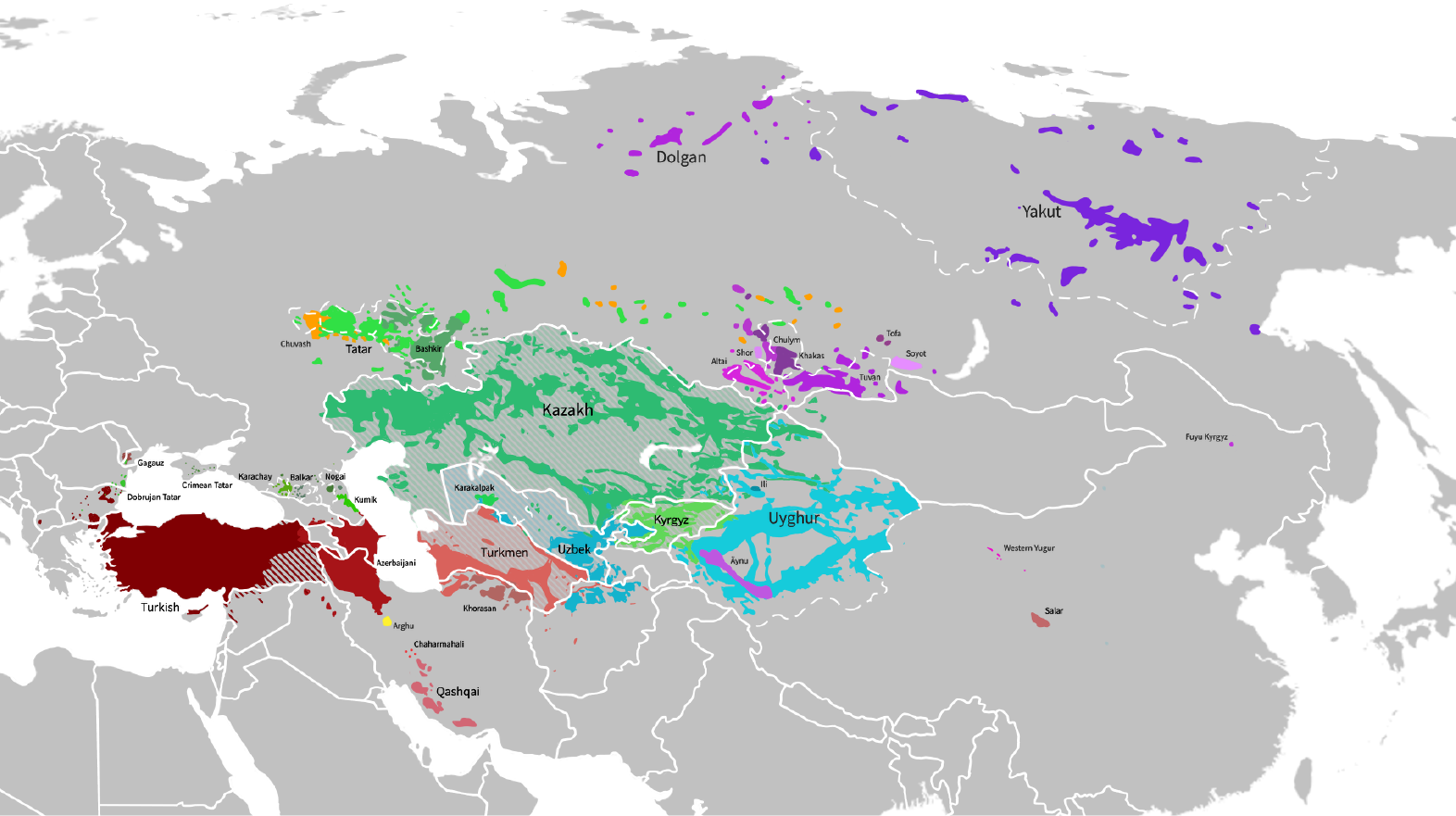}
    \caption{Geographic distribution of the Turkic language family (Source: Wikimedia Commons).}
    \label{fig:turkic_map}
\end{figure*}

The Turkic languages form one of the largest language families, with more than 35 languages spoken by about 200 million people from the Balkans and Anatolia to Siberia and north-west China (Figure~\ref{fig:turkic_map}; \citealt{johanson2021turkic}). They share agglutinative morphology, vowel harmony, and subject--object--verb word order, which distinguish them from many Indo-European languages \cite{jurafsky2023speech}, and they are closely related to each other: the sentence \emph{``Ali saw the book''}, for instance, is almost morpheme-for-morpheme parallel across the family (Table~\ref{tab:sov-sentence}). Orthographically, however, the family is fragmented. Turkic languages are written in Latin, Cyrillic, and Perso-Arabic scripts, several of them in more than one, and historical varieties use Perso-Arabic (Ottoman Turkish) or Old Turkic runes. Turkmenistan and Uzbekistan adopted the Latin script in 1993, and Kazakhstan plans to complete the change by 2031, while Cyrillic remains in wide use in all three countries. As a consequence, two closely related languages can look entirely unrelated to a model or tool that operates on surface strings.

\begin{table*}[!t]
\centering
\setlength{\tabcolsep}{4pt}
\small
\begin{tabular}{@{}p{2.85cm}p{4.55cm}@{\hspace{6pt}}p{2.85cm}p{4.55cm}@{}}
\toprule
\textbf{Language (ISO)} & \textbf{Sentence} &
\textbf{Language (ISO)} & \textbf{Sentence} \\
\midrule
\multicolumn{4}{l}{\textit{Latin script}} \\[1pt]
Turkish (tur)         & Ali kitab{\i} gördü.        & Azerbaijani (aze)    & Ali kitab{\i} gördü. \\
Turkmen (tuk)         & Ali kitaby gördi.           & Gagauz (gag)         & Ali kitab{\i} gördü. \\
Crimean Tatar (crh)   & Ali kitan{\i} kördi.        & Karakalpak (kaa)     & Ali kitapt{\i} kórdi. \\
Uzbek (uzb)           & Ali kitobni koʻrdi.         &                      & \\
\midrule
\multicolumn{4}{l}{\textit{Cyrillic script}} \\[1pt]
Kazakh (kaz)          & Әли кітапты көрді.          & Kyrgyz (kir)         & Али китепти көрдү. \\
Tatar (tat)           & Али китапны күрде.          & Bashkir (bak)        & Али китапты күрҙе. \\
Nogai (nog)           & Али китапты коьрди.         & Kumyk (kum)          & Али китапны гёрдю. \\
Karachay-Balkar (krc) & Али китапны кёрдю.          & Altai (alt)          & Али бичикти кӧрди. \\
Sakha (sah)           & Али кинигэни көрдө.         & Tuvan (tyv)          & Али номну көрген. \\
Chuvash (chv)         & Али кӗнекене курнӑ.         & Khakas (kjh)         & Али пічікті кӧрді. \\
\midrule
\multicolumn{4}{l}{\textit{Perso-Arabic script (right-to-left)}} \\[1pt]
Uyghur (uig) & ئەلى كىتابنى كۆردى. & & \\
\bottomrule
\end{tabular}
\caption{The sentence \emph{``Ali saw the book''} in 20 Turkic languages: subject, accusative-marked object, and sentence-final verb, written in the three active script families (Latin, Cyrillic, and Perso-Arabic).}
\label{tab:sov-sentence}
\end{table*}

\subsection{Motivation}
This combination of strong typological and lexical relatedness, weak orthographic relatedness, and highly uneven resources makes the Turkic family a natural test bed for multilingual and cross-lingual NLP. Existing tools, however, concentrate on Turkish and, increasingly, on Kazakh \cite{joshi2020state,oflazer2018turkish_nlp}. Turkish has mature toolkits such as Zemberek \cite{akin2007zemberek}, and Kazakh benefits from the resources released by ISSAI, but most other Turkic languages are served by isolated tools, by Apertium transducers that have no Python pipeline interface \cite{forcada2011apertium}, or not at all. Working with several Turkic languages therefore requires a separate toolchain for each language and script. The agglutinative morphology of the family further calls for morphological analysis beyond tokenization and tagging, and the ongoing script transitions call for reliable transliteration to connect data written in different orthographies. Finally, the subword tokenizers of large language models segment many Turkic languages into far more tokens than English (\S\ref{subsec:related-tokenizer}), which makes processing these languages slower and more expensive.

\subsection{Contributions}
In this paper, we present \textbf{TurkicNLP}, an open-source Python toolkit that brings these resources together in one pipeline for 20+ Turkic languages and four scripts. Our contributions are as follows:
\begin{itemize}
  \item \textbf{A unified pipeline with multiple backends.} A single \texttt{Pipeline} interface covers tokenization, multi-word token expansion, morphological analysis, POS tagging, lemmatization, dependency parsing, NER, sentence embeddings, and machine translation, with the same calling convention for every language. Each component can use a rule-based, finite-state (Apertium), per-language neural (Stanza), or multilingual neural (Glot500) backend, and documents are read and written in CoNLL-U.
  \item \textbf{Script-aware infrastructure for cross-lingual processing.} The toolkit provides automatic script detection, transliteration based on official standards for all ten languages that are written in more than one script, a conversion of 23 languages into the shared Latin orthography of the \emph{Common Turkic Alphabet}, language identification restricted to Turkic varieties, and transliteration bridging, which lets a model trained on one script process text in another.
  \item \textbf{Multilingual models with zero-shot transfer.} We train two Glot500-based models with script adapters and language embeddings. One provides POS tagging and dependency parsing for 15 languages, the other morphological analysis and lemmatization for 22 languages, including languages without annotated data, which are covered through the embedding of a related proxy language. We further combine finite-state and neural analyses in a hybrid morpheme segmenter.
  \item \textbf{An evaluation across the family.} We compare finite-state and neural morphological analysis on a parallel challenge set in 20 languages and per-language and multilingual parsers on Turkic UD test sets, evaluate transliteration, script detection, and language identification, and show that conversion to the Common Turkic Alphabet closes most of the script-driven tokenization gap of LLM tokenizers.
  \item \textbf{Open-source release.} The library is released under the Apache-2.0 license at \url{https://github.com/turkic-nlp/turkicnlp}, together with the evaluation code and companion Jupyter notebooks for all supported languages.\footnote{\url{https://github.com/turkic-nlp/turkic-nlp-code-samples}}
\end{itemize}

%% file: sections/02-related-work.tex
\section{Related Work}
\label{sec:related}

\subsection{General-Purpose NLP Toolkits}
\label{subsec:related-general}

General-purpose NLP toolkits have reached high maturity for major European languages but offer limited coverage of the Turkic family. Stanza \cite{qi2020stanza} provides neural pipeline components trained on Universal Dependencies treebanks for 66+ languages, including five Turkic languages: Turkish, Kazakh, Kyrgyz, Uyghur, and Ottoman Turkish. However, it lacks a morphology-first design, does not support script transliteration, and provides no integration with Apertium FSTs for the remaining Turkic languages. Trankit \cite{nguyen-etal-2021-trankit} is a transformer-based multilingual toolkit covering 56 languages via pipeline components built on XLM-R; it overlaps with Stanza on UD-treebank Turkic languages but similarly provides no script conversion or FST morphology. spaCy \cite{honnibal2020spacy} is an industrial-strength library optimized for efficiency across major European languages but offers no dedicated Turkic NLP support beyond standard Unicode handling. NLTK \cite{bird2009nltk} is a widely-used educational framework with minimal Turkic-specific components. Polyglot \cite{al-rfou-etal-2013-polyglot} provides distributed word representations and basic NLP components for 100+ languages, including several Turkic varieties, but does not support morphological analysis, script conversion, or unified pipeline integration. UDPipe \cite{straka2016udpipe} offers trainable pipelines for any CoNLL-U treebank but does not bundle Turkic models and lacks transliteration support. UralicNLP \cite{hamalainen2019uralicnlp}, built on the Giellatekno/Divvun FST infrastructure, provides a comparable morphology-first, script-aware toolkit for Uralic minority languages; its architecture---FST analyzers accessed through a unified Python API---directly parallels TurkicNLP's design, though it targets an entirely different language family with different morphological and script profiles. calamanCy \cite{miranda-2023-calamancy} and PyThaiNLP \cite{phatthiyaphaibun-etal-2023-pythainlp} exemplify a complementary paradigm: language-specific pipelines that consolidate disjoint NLP tools for morphologically complex under-resourced languages behind a unified API. TurkicNLP occupies a similar role---consolidating fragmented tools for under-resourced languages---but extends the paradigm to a 24-language family with integrated multi-backend swappability and script-aware transliteration routing as first-class architectural features.

\subsection{Turkic-Specific NLP Tools and Pipelines}
\label{subsec:related-turkic}

Several tools address individual Turkic languages, organized here by language.

\textbf{Turkish}: Oflazer and Sara\c{c}lar \cite{oflazer2018turkish_nlp} provide the definitive reference volume on Turkish NLP, surveying 25 years of work on morphological processing, language modeling, deep parsing, machine translation, treebanks, and lexical resources. Zemberek-NLP \cite{akin2007zemberek} is a comprehensive Java library for Turkish NLP covering morphological analysis, POS tagging, and spell checking; it is restricted to Turkish, does not expose a pip-installable Python API, and has no cross-family support. TurkishDelightNLP \cite{alecakir-etal-2022-turkishdelightnlp} is a neural Turkish NLP pipeline offering tokenization, morphological analysis, dependency parsing, and NER with strong performance on Turkish; it targets Turkish exclusively and does not integrate with other Turkic languages or script conversion. VNLP \cite{vnlp2023} is an open-source Python NLP library for Turkish providing morphological analysis, named entity recognition, dependency parsing, and sentiment analysis, including quantitative benchmarks on standard Turkish test sets; like TurkishDelightNLP it is Turkish-only and provides no script handling. Istanbul Technical University (ITU)\footnote{\url{https://bbf.itu.edu.tr/en/research/research-labs/nlp-lab}} maintains a complementary Turkish NLP infrastructure including neural pipelines, the IMST and ITU Web treebanks, TRopBank PropBank, and supporting corpora \cite{ddi_itu}. The Turkish NLP Suite \cite{turkish_nlp_suite} is an open-source non-profit collection that includes spaCy-compatible Turkish models, the SentiTurca sentiment corpus, TrGLUE benchmark, InstructTurca instruction-tuning data, and BellaTurca.

\textbf{Kazakh}: The Institute of Smart Systems and Artificial Intelligence (ISSAI) at Nazarbayev University\footnote{\url{https://issai.nu.edu.kz/}} has developed one of the most comprehensive Turkic-language NLP ecosystems. Text resources include KazNERD \cite{yeshpanov2022kaznerd} (112,702 tokens, 25 entity classes), KazQAD \cite{yeshpanov2024kazqad}, and KazParC \cite{yeshpanov-etal-2024-kazparc} (371,902 parallel sentences across Kazakh, English, Russian, and Turkish). The speech strand covers KazakhTTS2 \cite{mussakhojayeva-etal-2022-kazakhtts2} (271 hours TTS), a multilingual Turkic speech corpus spanning 10 languages \cite{mussakhojayeva2023multilingual}, and the Söyle noise-robust ASR system for 11 Turkic languages \cite{issai2023soyle}.

\textbf{Kyrgyz}: \citet{alekseev2025kyrgyznlp} survey the state of Kyrgyz NLP, cataloguing available resources and proposing a development roadmap; they note that despite millions of speakers, Kyrgyz remains severely under-resourced. KyrgyzNER \cite{turatali2025humanannotatednerdatasetkyrgyz} provides 10,900 annotated sentences across 27 entity classes for Kyrgyz NER. \citet{11206960} present a benchmark for evaluating Kyrgyz LLMs covering commonsense reasoning and multilingual understanding tasks. The Cramer Project \cite{cramerproject2024akylai} provides dedicated Kyrgyz speech models: AkylAI-STT-small (Whisper fine-tuned for Kyrgyz ASR) and AkylAI-TTS-mini (Matcha-TTS trained on 13 hours of Kyrgyz speech).

\textbf{Multiple languages}: Apertium \cite{forcada2011apertium} provides rule-based morphological analyzers and machine translation for 20+ Turkic languages via finite-state transducers, representing the most comprehensive freely available resource for Turkic morphological analysis. Its native interface is command-line-based, loading Apertium FST binaries via the \texttt{hfst} Python package without requiring system-level installation. \citet{washington2020multiscript} complement this with multi-script morphological transducers and transcribers for seven Turkic languages, directly addressing the script diversity that complicates cross-lingual NLP in the family. Abdurakhmonova et al. \citeyear{abdurakhmonova2022turkicnlptool} present an NLP tool for linguistic analysis of Turkic languages focusing on morphological annotation, though it addresses a limited language set without a Python package interface. Universal Dependencies treebanks for Turkic languages were created by dedicated community efforts: the Turkish IMST treebank \cite{sulubacak-etal-2016-universal}, the Tatar NMCTT treebank \cite{taguchi-etal-2022-universal}, the Uyghur UDT \cite{eli-etal-2016-universal}, the Yakut/Sakha treebank \cite{merzhevich-ferraz-gerardi-2022-introducing}, and Old Turkish \cite{derin-harada-2021-universal}, with a parallel aligned treebank for Azerbaijani, Kyrgyz, Turkish, and Uzbek \cite{eslami-etal-2025-parallel}. These contributions remain language-specific and are not integrated into a unified pipeline toolkit. \citet{mirzakhalov-etal-2021-large} provide the first large-scale MT study across 22 Turkic languages, contributing a parallel corpus of approximately 1.4 million sentences, bilingual baselines for 26 language pairs, multi-domain test sets, and human evaluation scores. \citet{mirzakhalov-etal-2021-evaluating} follow up with a multiway multilingual NMT evaluation across the same Turkic language set, analysing transfer between language pairs and identifying bottlenecks for lower-resource varieties. \citet{tukeyev2026turkicmt} extend this line of work by developing and cleaning parallel corpora for six Turkic languages (Kazakh, Azerbaijani, Kyrgyz, Turkish, Turkmen, and Uzbek) totalling over 3.8 million sentences, and fine-tuning NLLB-200 to achieve substantial gains in BLEU and chrF over the baseline for low-resource pairs.

\subsection{Multilingual Models and Benchmarks}
\label{subsec:related-multilingual}

Large-scale multilingual models and language-specific pretrained models now cover a growing number of Turkic languages, organized here by language.

\textbf{Turkish}: BERTurk \cite{schweter2020berturk} provides multiple Turkish BERT variants (cased, uncased, 128k vocabulary). TURNA \cite{uludogan-etal-2024-turna} provides a 1.1B-parameter Turkish encoder-decoder language model. Large generative LLMs include Kumru-2B \cite{vngrs2024kumru} and Trendyol-LLM \cite{trendyol2024llm}.

\textbf{Kazakh}: KAZ-LLM \cite{issai2024kazllm} (8B and 70B parameters, LLaMA 3.1-based) and Sherkala-Chat \cite{koto2025sherkala} are generative models targeting Kazakh.

\textbf{Azerbaijani}: aLLMA \cite{hajili2024allma} is a language model for Azerbaijani.

\textbf{Uzbek}: UzBERT \cite{mansurov2021uzbert}, TahrirchiBERT \cite{tahrirchi2021bert}, and BERTbek \cite{kuriyozov-etal-2024-bertbek} cover Uzbek.

\textbf{Kyrgyz}: KyrgyzBERT \cite{kyrgyzbert2023} is available for Kyrgyz.

\textbf{Multilingual}: Cross-lingual word embedding mappings for Turkic languages were studied by \citet{kuriyozov-etal-2020-cross}, who show that shared morphological structure enables effective alignment across Kazakh, Uzbek, Azerbaijani, and Turkish. Large-scale multilingual models such as mBERT \cite{devlin2019bert} and XLM-R \cite{conneau2020unsupervised} provide broad cross-lingual representations for Turkic NLP, but suffer from high tokenizer fertility on agglutinative languages and do not constitute processing pipelines. The mGPT multilingual generative model \cite{shliazhko-etal-2024-mgpt} covers nine Turkic languages (Azerbaijani, Bashkir, Chuvash, Kazakh, Kyrgyz, Tatar, Turkmen, Uzbek, and Sakha). The NLLB-200 model \cite{costa2022nllb} covers 200 languages including 14 Turkic variants for machine translation.

\textbf{Evaluation benchmarks}: Evaluation benchmarks for Turkic NLP have advanced considerably. TUMLU \cite{isbarov2025tumlu} covers 9 Turkic languages (Azerbaijani, Crimean Tatar, Turkish, Uyghur, Uzbek, Karakalpak, Kyrgyz, Kazakh, and Tatar) with 38K multiple-choice questions; Karde\c{s}-NLU \cite{senel2024kardes} benchmarks cross-lingual NLU transfer from Turkish into five related languages (Azerbaijani, Kazakh, Kyrgyz, Uzbek, and Uyghur); TR-MMLU \cite{bayram2025trmmlu} provides 6,200 multiple-choice questions across 62 subject categories specifically for benchmarking Turkish LLMs; TurkishMMLU \cite{yuksel2024turkishmmlu} offers over 10,000 curriculum-based multiple-choice questions across 9 subjects spanning sciences, mathematics, and culturally specific topics for Turkish; Cetvel \cite{abrek2025cetvel} is a unified 23-task evaluation suite assessing language understanding, generation, and cultural competence of LLMs in Turkish; and TurkBench \cite{toraman2026turkbench} provides a broad multi-task benchmark for evaluating Turkish LLMs across diverse NLP tasks. \citet{11206960} build a dedicated benchmark for Kyrgyz LLMs targeting commonsense reasoning and multilingual transfer. Veitsman and Hartmann \citeyear{veitsman-hartmann-2025-recent} survey the state of NLP for Central Asian Turkic languages (Kazakh, Kyrgyz, Uzbek, Uyghur, Turkmen), documenting persistent resource gaps and the challenges of cross-lingual transfer from higher-resource Turkic languages. Collectively, these benchmarks underscore the need for unified processing infrastructure to support training and evaluation pipelines.

\subsection{Speech Technologies for Turkic Languages}
\label{subsec:related-speech}

Speech processing resources for Turkic languages have expanded significantly through dedicated corpus construction and multilingual modeling efforts. ISSAI has led the development of both ASR and TTS systems spanning the Turkic family: TurkicASR \cite{mussakhojayeva2023multilingual} trains multilingual ASR models across 10 Turkic languages (Azerbaijani, Bashkir, Chuvash, Kazakh, Kyrgyz, Sakha, Tatar, Turkish, Uyghur, Uzbek), achieving a 56.7\% relative CER reduction through cross-lingual transfer compared to monolingual baselines. TurkicTTS \cite{yeshpanov2023turkictts} extends this multilingual paradigm to speech synthesis for a partially overlapping set of 10 languages (Azerbaijani, Bashkir, Kazakh, Kyrgyz, Sakha, Tatar, Turkish, Turkmen, Uyghur, Uzbek)---notably including Turkmen but not Chuvash. S\"oyle \cite{issai2023soyle} provides noise-robust ASR for 11 Turkic languages by fine-tuning Whisper with data augmentation.

Key speech corpora include KSC2 (1,200~hours of Kazakh; \citealt{mussakhojayeva2023turkksc}), TSC (218~hours of Turkish; \citealt{tsc2024turkish}), USC (105~hours of Uzbek; \citealt{usc2021}), TatSC (269~hours of Tatar; \citealt{tatsc2023tatar}), TatarTTS (an open-source TTS synthesis dataset for Tatar; \citealt{orel2024tatartts}), TurkmenSpeech (251~hours of Turkmen; \citealt{mamedov2024turkmenspeech}), and KazakhTTS2 (271~hours; \citealt{mussakhojayeva-etal-2022-kazakhtts2}).
Mozilla Common Voice \cite{ardila-etal-2020-common} provides community-recorded speech data for several Turkic languages, with Bashkir ($\sim$265~hours), Turkish ($\sim$129~hours), Kyrgyz ($\sim$44~hours), Tatar ($\sim$29~hours), Chuvash ($\sim$12~hours), Sakha ($\sim$6~hours), and Kazakh ($\sim$1~hour). FLEURS \cite{conneau2022fleurs} covers Azerbaijani, Kazakh, Kyrgyz, Turkish, and Uzbek for few-shot speech evaluation. Dedicated Kyrgyz speech models have also emerged: the Cramer Project \cite{cramerproject2024akylai} releases AkylAI-STT-small (Whisper fine-tuned for Kyrgyz) and AkylAI-TTS-mini (Matcha-TTS trained on 13~hours of Kyrgyz speech), providing the first openly available language-specific ASR and TTS systems for Kyrgyz. Beyond system building, several studies survey and extend ASR across the Uyghur--Kazakh--Kyrgyz cluster: \citet{du2023asroverview} review ASR techniques for Uyghur, Kazakh, and Kyrgyz, identifying shared phonological challenges and techniques effective across all three languages; \citet{orel2023speech} demonstrate cross-lingual transfer from Kazakh to other Turkic varieties using a shared-encoder approach; and \citet{bekarystankyzy2024multilingual} propose a multilingual end-to-end ASR system exploiting the common Cyrillic alphabet across low-resource Turkic languages.

\subsection{Tokenizer Efficiency Across Turkic Languages}
\label{subsec:related-tokenizer}

\begin{figure*}[t]
\centering
\begin{subfigure}[t]{\textwidth}
\centering
\includegraphics[width=\textwidth]{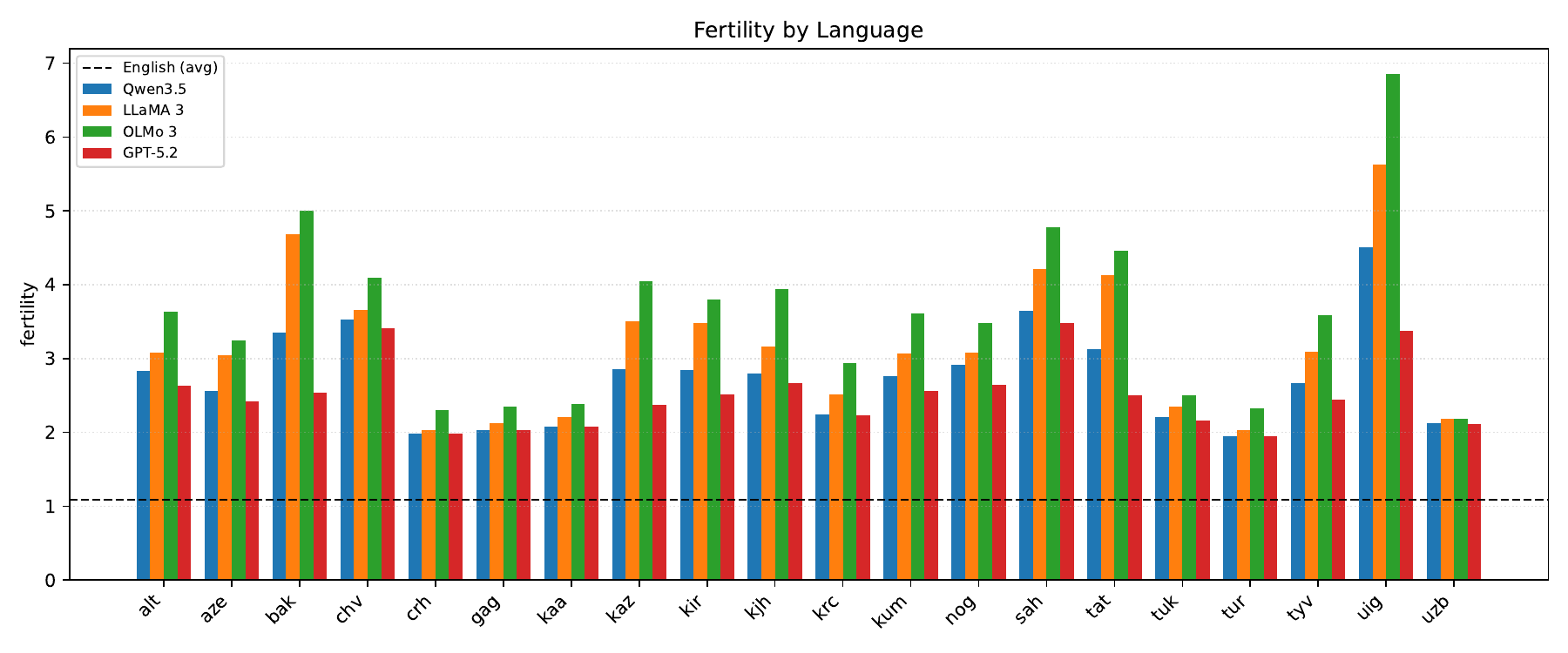}
\caption{Mean fertility (tokens per word).}
\label{fig:tokenizer-fertility}
\end{subfigure}

\vspace{0.5em}

\begin{subfigure}[t]{\textwidth}
\centering
\includegraphics[width=\textwidth]{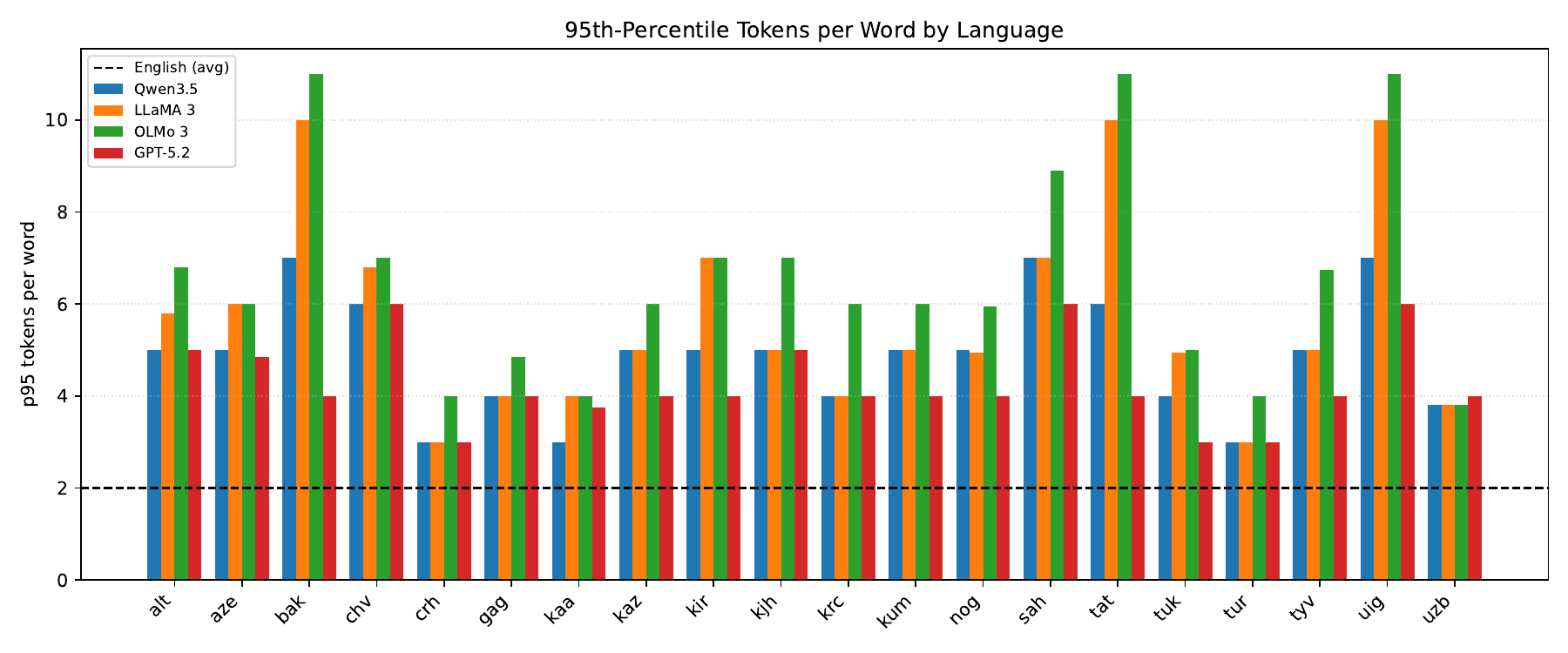}
\caption{95th-percentile tokens per word.}
\label{fig:tokenizer-p95}
\end{subfigure}
\caption{Tokenizer efficiency of four LLM tokenizers (Qwen3.5, LLaMA~3, OLMo~3, and GPT-5.2) on the 20 parallel challenge sentences in 20 Turkic languages and English. Higher values indicate less efficient segmentation.}
\label{fig:tokenizer-analysis}
\end{figure*}

The subword tokenizers of multilingual LLMs are trained on data dominated by a few high-resource languages, and long agglutinated Turkic words are therefore often split into many pieces. Fertility, the average number of tokens per word, has a direct effect on cost: API prices, compute, generation time, and context windows all scale with the number of tokens, so text with a fertility of 4 costs about twice as much as the same text with a fertility of 2, and only half as much of it fits into the context window. As a cross-lingual diagnostic, we measure the fertility of four widely used tokenizers, namely those of Qwen3.5-35B-A3B, Meta-LLaMA-3-8B, OLMo-3-1025-7B, and GPT-5.2 (tiktoken \texttt{o200k\_base}), on the 20 parallel challenge sentences in 20 Turkic languages that we also use for the morphological evaluation (\S\ref{subsec:eval-morph}, Appendix~\ref{sec:app-challenge}). Every word is tokenized separately and without special tokens.

Figure~\ref{fig:tokenizer-analysis} shows the results. All four tokenizers segment English at 1.08--1.09 tokens per word. Turkish, the best-represented Turkic language, needs between 1.95 (Qwen3.5 and GPT-5.2) and 2.33 (OLMo~3) tokens per word. For the less-resourced languages, the additional cost is considerably higher, although their morphology is comparable to that of Turkish: fertility ranges from 3.38 (GPT-5.2) to 6.85 (OLMo~3) for Uyghur, from 3.48 to 4.78 for Sakha, and from 2.54 to 5.00 for Bashkir. The 95th percentile of tokens per word, which approximates the worst-case segmentation of long agglutinated forms, reaches 11 for Uyghur and Tatar under OLMo~3 (10 under LLaMA~3), compared to 2 for English under all tokenizers. Averaged over the 20 languages, GPT-5.2 is the most efficient tokenizer (2.51), followed by Qwen3.5 (2.75), LLaMA~3 (3.16), and OLMo~3 (3.58).

The disparity follows the script more closely than the amount of resources or the linguistic distance. Perso-Arabic Uyghur and Cyrillic Bashkir, Sakha, Tatar, and Chuvash are affected most, whereas Latin-script languages fare best, including low-resource Crimean Tatar and Gagauz, whose fertility is on par with that of Turkish. The sample is small and intended as a diagnostic rather than a benchmark. Still, the pattern motivates two components of TurkicNLP: the transliteration layer, which can map all languages into a shared script (\S\ref{subsec:transliteration}; evaluated in \S\ref{subsec:eval-cts}), and the morpheme segmenter, which produces linguistically motivated subword units for training morphology-aware tokenizers (\S\ref{subsec:morph-seg}).

\subsection{Resource Collections and Community Initiatives}
\label{subsec:related-collections}

Several community-driven collections catalogue the growing body of Turkic NLP resources. The Turkish NLP resource list \cite{turkishnlp_list}\footnote{\url{https://github.com/maidis/nlp-tr}} is a community-maintained index of Turkish NLP tools, corpora, treebanks, and pre-trained models. The Turkic Interlingua (TIL) project \cite{til_github} \footnote{\url{https://github.com/turkic-interlingua/til-mt}} provides parallel corpora across 22 Turkic languages together with baseline systems for machine translation, automatic speech recognition, and text-to-speech, enabling systematic multilingual research across the family. The UD-Turkic research group\footnote{\url{https://github.com/ud-turkic}} works on harmonising Universal Dependencies treebanks across Turkic languages, producing the first fully aligned parallel UD treebanks for Azerbaijani, Kyrgyz, Turkish, and Uzbek \cite{eslami-etal-2025-parallel}. Organisational and workshop infrastructure complements these datasets: SIGTURK, the ACL Special Interest Group on Turkic NLP \cite{sigturk2026}\footnote{\url{https://sigturk.github.io}}, organises dedicated workshop venues that bring together researchers working on Turkic languages and provide an important venue for disseminating resources and evaluation results. The TurkLang conference series\footnote{\url{https://turklang.ieees.org/}}, held annually since 2013, serves as another key venue dedicated to computational processing of Turkic languages, fostering cross-institutional collaboration across the Turkic-speaking world. \citet{sulevmanov2020morpheme} introduce the Turkic Morpheme Web Portal, a collaborative platform aggregating morphological databases and linguistic tools across Turkic languages to support Turkology research. Building on this, \citet{gatiatullin2024turklang} present the Turklang linguistic knowledge graph, a unified semantic model integrating lexical, morphological, and syntactic resources for multiple Turkic languages into a queryable graph structure. \citet{cekinel2025turkicsurvey} provide a recent systematic survey of Turkic language modeling, reviewing available corpora, models, and benchmarks while highlighting persistent challenges around data scarcity, tokenization, and multilingual transfer across the family. \citet{sharipov2025uzbekturkmen} offer a complementary systematic review focused on Uzbek and Turkmen, cataloguing available NLP resources, methods, and benchmarks and documenting the particular challenges posed by agglutinative morphology and historical script transitions.

\subsection{Turkic NLP Resource Landscape}
\label{subsec:resource-landscape}

Table~\ref{tab:resources-full} (Appendix) catalogues datasets, NLP components, and toolkit coverage for all 24 languages, grouped by genetic branch. Table~\ref{tab:llm-models} (Appendix) surveys all publicly available pretrained neural language models---encoder-only, generative, and multilingual---for each supported language. Together, these tables motivate the design choices described in the subsequent sections.

%% file: sections/03-design.tex
\section{Library Design and Architecture}
\label{sec:design}

\subsection{Design Goals}
\label{subsec:design-goals}

TurkicNLP is designed around six principles. First, it offers a \textbf{unified interface}: a single \texttt{Pipeline} entry point works in the same way for all 24 languages, so that users can switch languages without changing their code. Second, it has \textbf{multiple backends}: each processor can use interchangeable rule-based, finite-state (Apertium), per-language neural (Stanza), or multilingual neural (Glot500) backends, selected through the catalog defaults or explicit options. Third, it is \textbf{script-aware}: input scripts are detected automatically, and text is transliterated into the script that a model was trained on and mapped back afterwards. Fourth, it has a \textbf{lightweight core} that depends only on NumPy and \texttt{regex} (Python $\geq$ 3.10), while heavier dependencies (PyTorch, Stanza, Hugging Face Transformers, HFST, fastText) are optional extras such as \texttt{pip install turkicnlp[hfst]}, \texttt{[stanza]}, \texttt{[transformers]}, \texttt{[translation]}, \texttt{[lid]}, or \texttt{[all]}. Fifth, licenses are kept \textbf{separate}: the library is Apache-2.0 licensed, while GPL-licensed Apertium transducers and CC-BY-NC NLLB weights are never bundled but downloaded at run time into separate cache directories. Sixth, documents can be read and written in \textbf{CoNLL-U}, which makes the toolkit compatible with the Universal Dependencies ecosystem \cite{demarneffe2021universal}.

\subsection{Pipeline Architecture}
\label{subsec:pipeline}

The \texttt{Pipeline} class is the central entry point. It takes a language code, a list of processors, an optional script, an optional bridge script (\texttt{transliterate\_to}), and per-processor options such as \texttt{morph\_backend="apertium"}. When a pipeline is built, it (1) determines the script that its models expect; (2) resolves processor dependencies, where the requirements are read from the backend that will actually run, so that requesting \texttt{depparse} adds \texttt{tokenize} and \texttt{pos}; (3) selects each backend from the user's choice or the catalog default for the language and script, never choosing an unfinished placeholder implementation; and (4) loads the processors in the following canonical order:

\begin{center}
\small
\texttt{tokenize $\to$ mwt $\to$ morph $\to$ morph\_neural $\to$ pos $\to$ feats $\to$ lemma $\to$ depparse $\to$ ner $\to$ embeddings $\to$ translate}
\end{center}

When several processors provide the same annotation field, the last one in this order owns it. A dedicated tagger therefore overrides the part of speech of a full morphological analyzer, while the analyzer's output is kept per word for inspection, and features are dropped for a word when they come from a processor whose own part of speech disagrees with the final one.

When the pipeline is applied to a text, it creates a \texttt{Document}, detects the script if none was declared, and, if the models of the language were trained on another script, transliterates the text into the model script. After all processors have run, the results are mapped back to the input, so that word forms and character offsets refer to the original text and lemmas are transliterated back. Empty input yields an empty document. Batches of texts are processed with \texttt{pipeline.batch(texts, batch\_size)}, and files with \texttt{pipeline.process\_file(input, output, format)} with \texttt{conllu} or \texttt{json} output. Figure~\ref{fig:architecture} shows the component structure of the library.

\begin{figure}[h]
\centering
\begin{tikzpicture}[
  font=\tiny,
  pipe/.style={draw=blue!55!black, fill=blue!7, rounded corners=3pt,
               align=center, inner sep=5pt, text width=3.2cm,
               minimum height=0.85cm},
  chain/.style={draw=orange!55!black, fill=orange!7, rounded corners=3pt,
                align=center, inner sep=5pt, text width=6.8cm},
  side/.style={draw=teal!55!black, fill=teal!7, rounded corners=3pt,
               align=center, inner sep=4pt, text width=1.55cm},
  docbox/.style={draw=violet!55!black, fill=violet!7, rounded corners=3pt,
                 align=center, inner sep=5pt, text width=4.5cm},
  arr/.style={->, >=Stealth, semithick}]

\node[side] (script) at (0.975, 0) {%
  {\bfseries Script}\\[2pt]
  ScriptDetector\\Transliterator\\LanguageDetection};

\node[pipe] (pipe) at (3.7, 0) {%
  {\bfseries\small Pipeline}\\[3pt]
  \texttt{lang} $\cdot$ \texttt{processors[]}\\
  \texttt{script} $\cdot$ \texttt{config}};

\node[side] (model) at (6.625, 0) {%
  {\bfseries Model}\\[1pt]{\bfseries Mgmt}\\[2pt]
  ModelRegistry\\Downloader\\\texttt{catalog.json}};

\draw[arr, <->] (script.east) -- (pipe.west);
\draw[arr, <->] (pipe.east) -- (model.west);

\node[chain] (chain) at (3.7, -2.25) {%
  {\bfseries Processor Chain}\\[4pt]
  \texttt{tokenize} $\to$ \texttt{mwt} $\to$ \texttt{morph} $\to$ \texttt{morph\_neural}
  $\to$ \texttt{pos} $\to$ \texttt{feats}\\[2pt]
  $\to$ \texttt{lemma}
  $\to$ \texttt{depparse} $\to$ \texttt{ner}
  $\to$ \texttt{embeddings} $\to$ \texttt{translate}};

\draw[arr] (pipe.south) -- (chain.north)
  node[midway, right=1pt, font=\tiny\itshape, text=gray] {orchestrates};

\node[docbox] (doc) at (3.7, -3.9) {%
  {\bfseries Document Model}\\[3pt]
  \texttt{Document} $\to$ \texttt{Sentence}\\
  $\hookrightarrow$ \texttt{Token} / \texttt{Word} / \texttt{Span}};

\draw[arr] (chain.south) -- (doc.north)
  node[midway, right=1pt, font=\tiny\itshape, text=gray] {produces};

\end{tikzpicture}
\caption{High-level component architecture of TurkicNLP. The \texttt{Pipeline} orchestrates a sequential \texttt{Processor} chain; the Script subsystem handles automatic script detection and transliteration; Model Management resolves and downloads model artifacts via \texttt{catalog.json}; the Document Model provides a hierarchical output representation.}
\label{fig:architecture}
\end{figure}

\subsection{Document Data Model}
\label{subsec:datamodel}

TurkicNLP uses a Stanza-compatible hierarchical data model:
\texttt{Document $\to$ Sentence $\to$ Token $\to$ Word $\to$ Span}.
Each \texttt{Word} carries: \texttt{text}, \texttt{lemma}, \texttt{upos} (Universal POS tag), \texttt{xpos} (language-specific tag), \texttt{feats} (UD morphological features), \texttt{head} and \texttt{deprel} (dependency arc), \texttt{deps} (enhanced dependencies), \texttt{start\_char}/\texttt{end\_char}, and \texttt{ner} label. Named entity spans are represented as \texttt{Span} objects with \texttt{label}, \texttt{start\_char}, and \texttt{end\_char}. The \texttt{Document} additionally records \texttt{script} (detected or declared), \texttt{embedding} (document-level vector from the NLLB backend), \texttt{translation} (output from the translation backend), and \texttt{\_processor\_log} tracing which backends processed the document. Documents round-trip through CoNLL-U, and \texttt{doc.to\_dict(full=True)} exports all word-, sentence-, and document-level annotations as JSON.

Listing~\ref{lst:docmodel} illustrates standard pipeline usage. CoNLL-U output is produced via \texttt{doc.to\_conllu()}.

\begin{lstlisting}[caption={Pipeline usage and annotation access (Turkish).},
                   label={lst:docmodel}]
import turkicnlp

turkicnlp.download("tur")

nlp = turkicnlp.Pipeline(
    "tur",
    processors=["tokenize","pos","lemma","ner","depparse"])

doc = nlp("Halil dün Ankara'ya gitti")

for w in doc.words:
    print(w.text, w.upos, w.lemma, w.ner)
# Halil      PROPN  Halil   O
# dün        NOUN   dün     O
# Ankara'ya  PROPN  Ankara  B-LOCATION
# gitti      VERB   git     O

print(doc.to_conllu())
\end{lstlisting}

\subsection{Processor Interface}
\label{subsec:processor-interface}

Each processor is a subclass of an abstract \texttt{Processor} base class that declares what annotation layers it consumes and produces. This declarative dependency model allows the pipeline to resolve processor ordering automatically and to raise an error early if a required layer is missing. The interface is intentionally minimal: processors implement a load method and a process method, and are otherwise self-contained. This design makes it straightforward for third-party contributors to add new languages, new backends, or new processing approaches without modifying the pipeline core.

\subsection{Model Registry and Download Manager}
\label{subsec:registry}

Model metadata is maintained in a central JSON manifest (\texttt{catalog.json}) that records, for each language, script, and processor, the available backends and the default one, and for each downloadable model its URL, SHA-256 checksum, license, and quality tier. On first use, the library downloads and verifies only the model files that the requested processors need; downloads are written to temporary files and renamed only when complete, so an interrupted download is never mistaken for a complete one, and shared models (e.g., the Glot500 backbone and NLLB) are stored only once. The catalog packaged with the library is used by default, and a user-supplied catalog can replace it, so that new models can be registered without a new release of the library.

%% file: sections/04-scripts.tex
\section{Script and Orthography Support}
\label{sec:scripts}

\subsection{Script Coverage}
\label{subsec:script-coverage}

TurkicNLP supports four script families: Latin (Latn), Cyrillic (Cyrl), Perso-Arabic (Arab), and Old Turkic runic (Orkh). Several languages are written in more than one script, and several countries are in the middle of an official script transition. Table~\ref{tab:coverage-matrix} lists the scripts of every language. Unicode NFC normalization is applied by default, and the Turkish dotted and dotless \textit{I} (\textit{\.{I}}/\textit{I}, \textit{i}/\textit{\i}) are handled correctly in lowercasing and uppercasing for the languages whose alphabets distinguish them. The pretrained neural language models available for each language are surveyed in Table~\ref{tab:llm-models} (Appendix).

\subsection{Automatic Script Detection}
\label{subsec:script-detection}

The script detector identifies the dominant script of a text by assigning each letter to the Unicode ranges of the four supported scripts; digits, punctuation, and whitespace are ignored. Cyrillic and Latin homoglyphs (e.g., Cyrillic \textit{а}, \textit{е}, \textit{о}) are classified by their code point rather than their appearance. A segmentation function additionally splits mixed-script text into runs of a single script, and an optional check compares the detected script with the scripts of a declared language. For text of unknown language, \texttt{LanguageDetection} wraps GlotLID \cite{kargaran-etal-2023-glotlid}, whose label space is restricted by default to the Turkic language--script labels of the toolkit.

\subsection{Transliteration}
\label{subsec:transliteration}

TurkicNLP implements transliteration for all ten languages that are written in more than one script (Table~\ref{tab:translit-pairs}). Conversion between Cyrillic and Latin works in both directions and follows the official standards for Kazakh (2021), Uzbek (1995), Azerbaijani (1991), Turkmen (1993, revised), Karakalpak (2016), Tatar (Zamanälif), and Crimean Tatar. Uyghur is converted among Perso-Arabic, Latin (Uyghur Latin Yéziqi), and Cyrillic with an integrated multi-script converter \cite{uyghur_transliteration}. Ottoman Turkish (Latin to Perso-Arabic, following a simplified academic transcription) and Old Turkic (runic to Latin, covering the Orkhon and Yenisei letters of the Unicode block following Turkological conventions) are converted in one direction only.

\paragraph{Common Turkic Alphabet.} Every language except Old Turkic can also be converted to the Common Turkic Alphabet (CTS), a Latin alphabet of 34 letters designed to facilitate communication among Turkic-speaking nations \cite{turkic_academy_2024}. It was originally proposed at the 1991 Istanbul symposium and standardized by the Organization of Turkic States in 2024 \cite{istanbul_symposium_1991,ots_alphabet_2024}. CTS aims at one letter per phoneme across the family, so that cognates written differently in the national scripts converge (Kazakh \textit{Қазақстан} $\to$ \textit{Qazaqstan}, Azerbaijani \textit{Azərbaycan} $\to$ \textit{Azärbaycan}). This gives 23 languages a shared surface form for cross-lingual comparison and for models with a shared vocabulary; we evaluate its effect on LLM tokenization in \S\ref{subsec:eval-cts}.

\begin{table}[h]
  \centering
  \footnotesize
  \setlength{\tabcolsep}{1.5pt}
  \begin{tabular}{llll}
    \toprule
    \textbf{Language} & \textbf{ISO} & \textbf{Direction} & \textbf{Standard} \\
    \midrule
    Kazakh          & kaz & Cyrl $\leftrightarrow$ Latn & Official 2021 \\
    Uzbek           & uzb & Cyrl $\leftrightarrow$ Latn & Official 1995 \\
    Azerbaijani     & aze & Cyrl $\leftrightarrow$ Latn & Official 1991 \\
    Tatar           & tat & Cyrl $\leftrightarrow$ Latn & Zamanälif \\
    Turkmen         & tuk & Cyrl $\leftrightarrow$ Latn & 1993 (revised) \\
    Karakalpak      & kaa & Cyrl $\leftrightarrow$ Latn & Official 2016 \\
    Crimean Tatar   & crh & Cyrl $\leftrightarrow$ Latn & Latin std. \\
    Uyghur          & uig & Arab $\leftrightarrow$ Latn $\leftrightarrow$ Cyrl & ULY / UKY \\
    Ottoman Turkish & ota & Latn $\to$ Arab & Academic \\
    Old Turkic      & otk & Orkh $\to$ Latn & Turkological \\ \hline
    23 languages    & --  & native $\to$ CTS & OTS 2024 \\

    \bottomrule
  \end{tabular}
  \caption{Transliteration pairs supported by TurkicNLP. Uyghur additionally supports conversion from CTS back to the native scripts.}
  \label{tab:translit-pairs}
\end{table}

The engine applies greedy longest-match over mapping tables: at each position, it tries matches of decreasing length (up to four characters), so that digraphs and trigraphs are resolved before single characters. This is complemented by context rules for letters whose value depends on their position, e.g., \textit{е}/\textit{э} versus \textit{e}/\textit{ye} in Uzbek, Turkmen, and Tatar, Kazakh \textit{и} and \textit{й}, which are both written \textit{i} in the 2021 alphabet, Uzbek \textit{сҳ}, which becomes \textit{sʼh} so that it is not read as \textit{sh}, and Crimean Tatar \textit{ё} and \textit{ю}, which stand for \textit{ö} and \textit{ü} after consonants. Capitalization is preserved, including words written entirely in capitals. The pipeline integrates transliteration automatically (\S\ref{subsec:pipeline}): when the models of a language were trained on another script than that of the input, the text is converted to the model script and the results are mapped back, which makes script conversion transparent to the user. Known limitations include word-initial \textit{ö} and \textit{ü} in Crimean Tatar, which Cyrillic writes as plain \textit{о} and \textit{у}; Russian loanwords with lexical exceptions in Uzbek Cyrillic (e.g., \textit{film} $\to$ \textit{фильм}); and Ottoman Turkish transcription diacritics such as \textit{â} or \textit{ḥ}, which are not yet mapped. We evaluate transliteration in \S\ref{subsec:eval-scripts}.

\begin{lstlisting}[caption={Script detection and Kazakh Cyrillic/Latin transliteration.},
                   label={lst:translit}]
from turkicnlp.scripts import Script
from turkicnlp.scripts.detector import detect_script
from turkicnlp.scripts.transliterator import Transliterator

# Detect script
cyrl = "Мен Алматыда турамын."  # Kazakh Cyrillic
print(detect_script(cyrl))    # Script.CYRILLIC

# Cyrillic -> Latin (2021 official Kazakh alphabet)
t = Transliterator(
    "kaz", source=Script.CYRILLIC, target=Script.LATIN)
latin = t.transliterate(cyrl)
print(latin)   # Men Almatyda turamyn.

# Reverse: Latin -> Cyrillic
t_back = Transliterator(
    "kaz", source=Script.LATIN, target=Script.CYRILLIC)
print(t_back.transliterate(latin))
\end{lstlisting}

\begin{lstlisting}[caption={Transliteration to the Common Turkic Alphabet.},
                   label={lst:common_translit}]
from turkicnlp.scripts import Script
from turkicnlp.scripts.transliterator import Transliterator

# Turkish Latin → CTS
t = Transliterator("tur", Script.LATIN, Script.COMMON_TURKIC)
print(t.transliterate("Türkçe güzel bir dil"))
# → Türkçe güzel bir dil

# Kazakh Cyrl → CTS
t = Transliterator("kaz", Script.CYRILLIC, Script.COMMON_TURKIC)
print(t.transliterate("Қазақстан"))
# → Qazaqstan

# Azerbaijani Latin → CTS
t = Transliterator("aze", Script.LATIN, Script.COMMON_TURKIC)
print(t.transliterate("Azərbaycan dili"))
# → Azärbaycan dili

# Turkmen Latin → CTS
t = Transliterator("tuk", Script.LATIN, Script.COMMON_TURKIC)
print(t.transliterate("Türkmen dili baýdyr"))
# → Türkmen dili baydır
\end{lstlisting}

\subsection{Language and Processor Coverage}
\label{subsec:coverage-matrix}

Table~\ref{tab:coverage-matrix} summarizes the scripts and the components available for each of the 24 supported languages. The coverage mirrors the uneven resource landscape. Turkish and Kazakh have the full pipeline including NER; ten languages have per-language neural pipelines (five with official Stanza models and five with our custom-trained models); 20 languages have finite-state morphology; and the multilingual Glot500 models extend tagging and parsing to 15 languages and morphological analysis and lemmatization to 22 languages, including languages with little or no annotated data. NLLB-200 provides sentence embeddings and machine translation for the eleven Turkic languages it covers.

\begin{table*}[!t]
  \centering
  \footnotesize
  \setlength{\tabcolsep}{4pt}
  \renewcommand{\arraystretch}{0.92}
  \begin{tabular}{@{}p{2.4cm}p{2.2cm}cccccccccccc@{}}
    \toprule
    \textbf{Language} & \textbf{Script(s)} &
    \rotatebox{90}{\textbf{Tokenize}} &
    \rotatebox{90}{\textbf{MWT}} &
    \rotatebox{90}{\textbf{Morph (FST)}} &
    \rotatebox{90}{\textbf{Morph (Neu)}} &
    \rotatebox{90}{\textbf{POS Tag}} &
    \rotatebox{90}{\textbf{Lemmatize}} &
    \rotatebox{90}{\textbf{Dep.\ Parse}} &
    \rotatebox{90}{\textbf{NER}} &
    \rotatebox{90}{\textbf{Translit.}} &
    \rotatebox{90}{\textbf{CTS}} &
    \rotatebox{90}{\textbf{Morph.\ Seg.}} &
    \rotatebox{90}{\textbf{Embed./MT}} \\
    \midrule
    \multicolumn{14}{@{}l}{\textbf{Oghuz}} \\
    \midrule
    Turkish          & Latn           & \tklyes & \tklyes & \tklyes & \tklyes & \tklyes & \tklyes & \tklyes & \tklyes & --      & \tklyes & \tklpar & \tklyes \\
    Azerbaijani      & Latn/Cyrl      & \tklyes & \tklno  & \tklyes & \tklyes & \tklyes & \tklyes & \tklyes & \tklno  & \tklyes & \tklyes & \tklpar & \tklyes \\
    South Azerb.     & Arab           & \tklyes & \tklno  & \tklno  & \tklno  & \tklno  & \tklno  & \tklno  & \tklno  & --      & \tklyes & \tklno  & \tklyes \\
    Turkmen          & Latn/Cyrl      & \tklyes & \tklno  & \tklpar & \tklyes & \tklyes & \tklyes & \tklyes & \tklno  & \tklyes & \tklyes & \tklpar & \tklyes \\
    Gagauz           & Latn           & \tklyes & \tklno  & \tklpar & \tklpar & \tklpar & \tklpar & \tklno  & \tklno  & --      & \tklyes & \tklpar & \tklno  \\
    \midrule
    \multicolumn{14}{@{}l}{\textbf{Kipchak}} \\
    \midrule
    Kazakh           & Cyrl/Latn      & \tklyes & \tklyes & \tklyes & \tklyes & \tklyes & \tklyes & \tklyes & \tklyes & \tklyes & \tklyes & \tklpar & \tklyes \\
    Kyrgyz           & Cyrl           & \tklyes & \tklyes & \tklyes & \tklyes & \tklyes & \tklyes & \tklyes & \tklno  & --      & \tklyes & \tklpar & \tklyes \\
    Tatar            & Cyrl/Latn      & \tklyes & \tklno  & \tklyes & \tklyes & \tklyes & \tklyes & \tklyes & \tklno  & \tklyes & \tklyes & \tklpar & \tklyes \\
    Bashkir          & Cyrl           & \tklyes & \tklno  & \tklpar & \tklyes & \tklyes & \tklyes & \tklyes & \tklno  & --      & \tklyes & \tklpar & \tklyes \\
    Crimean Tatar    & Latn/Cyrl      & \tklyes & \tklno  & \tklpar & \tklpar & \tklpar & \tklpar & \tklno  & \tklno  & \tklyes & \tklyes & \tklpar & \tklyes \\
    Karakalpak       & Latn/Cyrl      & \tklyes & \tklno  & \tklpar & \tklpar & \tklpar & \tklpar & \tklpar & \tklno  & \tklyes & \tklyes & \tklpar & \tklno  \\
    Nogai            & Cyrl           & \tklyes & \tklno  & \tklpar & \tklpar & \tklpar & \tklpar & \tklpar & \tklno  & --      & \tklyes & \tklpar & \tklno  \\
    Kumyk            & Cyrl           & \tklyes & \tklno  & \tklpar & \tklpar & \tklpar & \tklpar & \tklpar & \tklno  & --      & \tklyes & \tklpar & \tklno  \\
    Karach.-Balkar   & Cyrl           & \tklyes & \tklno  & \tklpar & \tklpar & \tklpar & \tklpar & \tklpar & \tklno  & --      & \tklyes & \tklpar & \tklno  \\
    \midrule
    \multicolumn{14}{@{}l}{\textbf{Karluk}} \\
    \midrule
    Uzbek            & Latn/Cyrl      & \tklyes & \tklno  & \tklyes & \tklyes & \tklyes & \tklyes & \tklyes & \tklno  & \tklyes & \tklyes & \tklpar & \tklyes \\
    Uyghur           & Arab/Latn/Cyrl & \tklyes & \tklno  & \tklpar & \tklyes & \tklyes & \tklyes & \tklyes & \tklno  & \tklyes & \tklyes & \tklpar & \tklyes \\
    \midrule
    \multicolumn{14}{@{}l}{\textbf{Siberian}} \\
    \midrule
    Sakha            & Cyrl           & \tklyes & \tklno  & \tklpar & \tklpar & \tklpar & \tklpar & \tklpar & \tklno  & --      & \tklyes & \tklpar & \tklno  \\
    Altai            & Cyrl           & \tklyes & \tklno  & \tklpar & \tklpar & \tklpar & \tklpar & \tklno  & \tklno  & --      & \tklyes & \tklpar & \tklno  \\
    Tuvan            & Cyrl           & \tklyes & \tklno  & \tklpar & \tklpar & \tklpar & \tklpar & \tklno  & \tklno  & --      & \tklyes & \tklpar & \tklno  \\
    Khakas           & Cyrl           & \tklyes & \tklno  & \tklpar & \tklpar & \tklpar & \tklpar & \tklno  & \tklno  & --      & \tklyes & \tklpar & \tklno  \\
    \midrule
    \multicolumn{14}{@{}l}{\textbf{Oghur}} \\
    \midrule
    Chuvash          & Cyrl           & \tklyes & \tklno  & \tklpar & \tklpar & \tklpar & \tklpar & \tklno  & \tklno  & --      & \tklyes & \tklpar & \tklno  \\
    \midrule
    \multicolumn{14}{@{}l}{\textbf{Arghu}} \\
    \midrule
    Khalaj           & Latn           & \tklyes & \tklno  & \tklno  & \tklpar & \tklpar & \tklpar & \tklno  & \tklno  & --      & \tklyes & \tklpar & \tklno  \\
    \midrule
    \multicolumn{14}{@{}l}{\textbf{Historical}} \\
    \midrule
    Ottoman Turk.    & Arab/Latn      & \tklyes & \tklyes & \tklno  & \tklyes & \tklyes & \tklyes & \tklyes & \tklno  & \tklyes & \tklyes & \tklpar & \tklno  \\
    Old Turkic       & Latn/Orkh      & \tklyes & \tklno  & \tklno  & \tklno  & \tklno  & \tklno  & \tklno  & \tklno  & \tklyes & \tklno  & \tklno  & \tklno  \\
    \midrule
    \textbf{\#\,languages} & & 24 & 4 & 20 & 22 & 22 & 22 & 15 & 2 & 10 & 23 & 22 & 11 \\
    \bottomrule
  \end{tabular}
  \caption{Component availability for the 24 languages of TurkicNLP (primary script first). \tklyes~=~production/stable; \tklpar~=~beta, prototype, paradigm-only training, or zero-shot; \tklno~=~not available; --~=~not applicable (single-script language). Morph (Neu): Glot500 morphological analyzer; POS, Lemmatize, and Dep.\ Parse combine Stanza and Glot500 backends (without a tagger, UPOS comes from the morphological analyzer). Translit.: native-script conversion; CTS: Common Turkic Alphabet; Morph.\ Seg.: hybrid morpheme segmenter; Embed./MT: NLLB-200.}
  \label{tab:coverage-matrix}
\end{table*}

%% file: sections/05-processors.tex
\section{Processing Capabilities}
\label{sec:processors}

The processing pipeline maps each stage to one or more interchangeable backends. Tokenization uses rule-based tokenizers for Latin/Cyrillic and Arabic-script text and neural Stanza \cite{qi2020stanza} tokenizers for the ten languages with Stanza models. Morphological analysis is provided by Apertium transducers \cite{forcada2011apertium} for 20 languages: Kyrgyz~\cite{washington-etal-2012-kyrgyz}, Tatar and Bashkir~\cite{tyers-etal-2012-tatar}, Kazakh and Tatar~\cite{salimzyanov-etal-2013-kazakh}, Kazakh, Tatar, and Kumyk~\cite{washington-etal-2014-kazakh}, Tuvan~\cite{tyers-etal-2016-tuvan,washington-etal-2016-tuvan-syntax}, Kazakh and Turkish~\cite{bayatli-etal-2018-kazakh}, Crimean Tatar~\cite{tyers-etal-2019-crimean}, Crimean Tatar and Turkish~\cite{gokirmak-etal-2019-crimean}, Crimean Tatar, Kazakh, Kyrgyz, Turkmen, Karakalpak, Uzbek, and Uyghur~\cite{washington-etal-2020-multiscript}, and Sakha~\cite{ivanova-etal-2022-sakha}, and by a multilingual neural analyzer for 22 languages. POS tagging, lemmatization, and dependency parsing are provided by Stanza models for ten languages and by a multilingual Glot500 \cite{imanigooghari-etal-2023-glot500} parser for 15 languages. Named entity recognition is available for Turkish and Kazakh, and sentence embeddings and machine translation use NLLB-200 \cite{costa2022nllb} for the eleven Turkic languages it covers. Table~\ref{tab:coverage-matrix} summarizes the availability of all components; runnable code samples for all languages are available in the companion repository.\footnote{\url{https://github.com/turkic-nlp/turkic-nlp-code-samples}} The following subsections describe each stage in detail.

\subsection{Tokenization and Multi-Word Token Expansion}
\label{subsec:tokenization}

TurkicNLP provides three tokenization backends. A rule-based tokenizer handles Latin- and Cyrillic-script input. It splits sentences at terminal punctuation, separates punctuation, brackets, and quotation marks (including \textit{«»}) from words, keeps apostrophes inside words (e.g., Turkish \textit{Ankara'ya}, Uzbek \textit{o‘}), and keeps abbreviations (e.g., \textit{Prof.}, \textit{т.б.}), ordinal numbers (\textit{1.}), decimals, and ellipses intact. A separate tokenizer for Arabic-script text handles right-to-left text, Arabic punctuation and decimals, and zero-width non-joiners in Uyghur and Ottoman Turkish. Both rule-based tokenizers compute character offsets on the original text. For nine of the ten languages with Stanza models, the neural Stanza tokenizer, trained on UD data, is the default: Turkish, Kazakh, Kyrgyz, and Uyghur with official models, and Uzbek, Azerbaijani, Turkmen, Tatar, and Bashkir with our custom-trained models; Ottoman Turkish uses the Arabic-script tokenizer. Multi-word tokens are expanded either by Stanza or by rule tables that we extracted from the UD treebanks of Turkish (3.9K rules), Ottoman Turkish, Kazakh, and Kyrgyz; every rule is checked so that its parts reconstruct the surface form.

\subsection{Morphological Analysis}
\label{subsec:morphology}

Morphological analysis is provided through Apertium finite-state transducer (FST) binaries compiled for 20 Turkic languages, accessed via Python-native HFST bindings without requiring a system-level Apertium installation. For a given token, the FST returns zero or more analyses—each comprising a lemma and a morphological tag string—which a tag-mapping subsystem converts to Universal Dependencies features and UPOS labels following the conventions established by \citet{tyers-washington-2015-kazakh}. Language-specific mappers exist for 21 languages, and the remaining languages use a common Turkic mapping; tags whose meaning depends on the part of speech (e.g., \texttt{<ind>}, indicative on verbs but indefinite on pronouns) are mapped accordingly. Morphological ambiguity is resolved using a combination of FST-internal HFST weights and heuristic scoring based on syntactic position, POS category priority, and lookups in curated closed-class lexicons for 21 languages; an external Apertium tagger can be configured as an additional disambiguation step. This remains a known limitation for agglutinative languages where local context strongly determines the correct reading; a lightweight sequence-based disambiguator is planned for a future release. The FST analyzers additionally support \textit{generation mode}: given a lemma and feature string, the transducer synthesizes the corresponding surface form, useful for data augmentation tasks. Table~\ref{tab:apertium-quality} lists quality levels per language.

\begin{table}[h]
  \centering
  \footnotesize
  \begin{tabular}{llll}
    \toprule
    \textbf{Language} & \textbf{ISO} & \textbf{Branch} & \textbf{Quality} \\
    \midrule
    Turkish          & tur & Oghuz    & Production \\
    Kazakh           & kaz & Kipchak  & Production \\
    Tatar            & tat & Kipchak  & Production \\
    Azerbaijani      & aze & Oghuz    & Stable \\
    Kyrgyz           & kir & Kipchak  & Stable \\
    Uzbek            & uzb & Karluk   & Stable \\
    Turkmen          & tuk & Oghuz    & Beta \\
    Bashkir          & bak & Kipchak  & Beta \\
    Uyghur           & uig & Karluk   & Beta \\
    Crimean Tatar    & crh & Kipchak  & Beta \\
    Chuvash          & chv & Oghur    & Beta \\
    Sakha            & sah & Siberian & Prototype \\
    Karakalpak       & kaa & Kipchak  & Prototype \\
    Nogai            & nog & Kipchak  & Prototype \\
    Kumyk            & kum & Kipchak  & Prototype \\
    Karachay-Balkar  & krc & Kipchak  & Prototype \\
    Altai            & alt & Siberian & Prototype \\
    Tuvan            & tyv & Siberian & Prototype \\
    Khakas           & kjh & Siberian & Prototype \\
    Gagauz           & gag & Oghuz    & Prototype \\
    \bottomrule
  \end{tabular}
  \caption{Apertium FST analyzer quality levels (based on Apertium lexicon size and UD tag-mapping completeness). \textit{Production}: estimated $>$85\% token coverage, full UD FEATS mapping. \textit{Stable}: $>$70\% coverage, partial UD mapping. \textit{Beta}: functional with known lexicon gaps. \textit{Prototype}: limited coverage, Apertium-native tags only.}
  \label{tab:apertium-quality}
\end{table}

\subsection{POS Tagging, Lemmatization, and Dependency Parsing}
\label{subsec:pos-dep}

For the five Turkic languages with UD treebanks, TurkicNLP provides neural POS tagging, lemmatization, and dependency parsing via Stanza \cite{qi2020stanza}, producing UPOS, XPOS, morphological features, lemma, and CoNLL-U dependency arcs in a single pass using biaffine attention-based models. Turkish offers seven treebank variants: IMST (general-domain default), BOUN, FrameNet, KeNet, ATIS, Penn, and Tourism; Kazakh uses KTB, Kyrgyz the KTMU treebank, Uyghur the UDT treebank, and Ottoman Turkish the BOUN treebank. The active treebank may be specified as a pipeline configuration parameter. In addition, custom-trained Stanza models provide full neural pipeline support (tokenization, POS tagging, lemmatization, and dependency parsing) for Uzbek, Azerbaijani, Turkmen, Tatar, and Bashkir, extending neural parsing coverage to ten Turkic languages in total. For these five languages, we used existing UD data and, where no such data was available, silver treebanks that we created by translating Turkish UD sentences with an LLM and annotating and validating the translations automatically. For Apertium-backed morphological analysis on any of the 20 supported languages, the same \texttt{Pipeline} interface applies with \texttt{morph\_backend="apertium"} and \texttt{processors=["tokenize","morph"]}; the pipeline then populates \texttt{w.lemma} and \texttt{w.feats} from the FST output (see Listing~\ref{lst:docmodel} for the analogous Turkish example).

\subsection{Multilingual Neural Models (Glot500)}
\label{subsec:glot500-models}

In addition to the per-language Stanza models, TurkicNLP provides two multilingual neural models with a common architecture: a Glot500 \cite{imanigooghari-etal-2023-glot500} encoder, followed by trainable script-specific adapters (Latin, Cyrillic, Arabic), language embeddings, and a shared two-layer BiLSTM, which feeds into task-specific heads. This design forces cross-lingual transfer through shared weights while accommodating the script diversity of the family through lightweight adapters. Sentences longer than the 512-subword limit of the encoder are processed in windows of whole words, and both models share one backbone in memory.

\paragraph{POS tagger and dependency parser.}
The first model keeps the whole Glot500 encoder frozen and jointly predicts UPOS tags and labeled dependency arcs with a biaffine dependency head \cite{dozat2017deep}; trees are decoded with the Chu-Liu/Edmonds algorithm, so that every sentence receives a well-formed tree with a single root. The model is trained on 20 UD-format treebanks of ten Turkic languages (116K sentences): Turkish (9 treebanks, 82K sentences), Azerbaijani, Turkmen, Bashkir, Uyghur, Kyrgyz, Ottoman Turkish, Tatar, Uzbek, and Kazakh, where the Azerbaijani, Turkmen, and Bashkir treebanks are silver (\S\ref{subsec:pos-dep}). Five more languages are covered zero-shot through the embedding of a related proxy language: Karakalpak via Uzbek; Kumyk, Karachay-Balkar, and Nogai via Tatar; and Sakha via Bashkir. A single model thus parses 15 languages, and the two parsing backends are interchangeable through \texttt{pos\_backend} and \texttt{depparse\_backend}.

\paragraph{Morphological analyzer and lemmatizer.}
The second model predicts UPOS tags, 69 UD feature--value pairs (as multi-label sigmoid classification), and lemmas, which are predicted as edit scripts by a character-level CNN. It was trained in two phases, first with the backbone frozen and then with its top three layers fine-tuned with discriminative learning rates. The training data combines three sources: (1) the UD-format treebanks of the same ten languages, which provide words in sentence context; (2) UniMorph paradigm tables \cite{sylak-glassman-2016-unimorph} for 13 Turkic languages (about 1.9M entries); and (3) inflectional paradigms extracted from Wiktionary with Wiktextract \cite{ylonen-2022-wiktextract} for 20 languages (about 4.8M forms). Languages without a treebank use the embedding of a related proxy language, which we choose by genealogical and areal proximity: Turkish for Crimean Tatar, Gagauz, and Khalaj; Bashkir for Sakha, Tuvan, Khakas, Altai, and Chuvash; and Tatar for Kumyk. For these languages, the paradigm data adapts the shared layers, whereas Karakalpak (via Uzbek), Karachay-Balkar, and Nogai (via Tatar) are covered purely zero-shot. In total, the model serves 22 languages of the toolkit (23 model languages, as Southern and Northern Altai both map to Altai). Lemmas are lowercased with Turkic-aware rules (\textit{I}~$\to$~\textit{ı} only in languages with a dotless \textit{ı}), and proper nouns keep their initial capital.

Both models use temperature sampling ($T{=}2$) during training to balance the heavily skewed data distribution (82K Turkish vs.\ 1K Kazakh sentences) and are installed with \texttt{pip install turkicnlp[transformers]}.

\subsection{Morpheme Segmentation}
\label{subsec:morph-seg}
The \texttt{MorphemeTokenizer} splits inflected words into labeled surface morphemes, e.g., Kazakh \textit{бармадым} `I did not go' $\to$ \textit{бар}/\textsc{stem} \textit{ма}/\textsc{neg} \textit{ды}/\textsc{pst} \textit{м}/\textsc{1sg}. Each Apertium reading yields a candidate segmentation, with the lemma as the stem and the tags as suffixes, and the candidate that accounts for the whole word is kept. The neural analyzer breaks ties and handles words unknown to the transducer, and derivational suffixes are split off only when the remaining base is a known word. To align the tags with the surface string, we use tables of suffix allomorphs for 23 languages that encode vowel harmony, consonant assimilation, and stem alternations and that combine curated rules with rules extracted from Wiktionary. A corpus annotator applies the segmenter at scale to produce training data for morphology-aware SentencePiece tokenizers, which are available as the \texttt{sp\_tokenize} processor. We evaluate the segmenter in \S\ref{subsec:eval-morph} and Appendix~\ref{sec:app-morphseg}.

\subsection{Named Entity Recognition}
\label{subsec:ner}

NER is available at production quality for Turkish and Kazakh. The Turkish model, integrated via Stanza, is trained on the Starlang annotated corpus and recognizes four entity types (person, organization, location, miscellaneous). The Kazakh model is trained on KazNERD \cite{yeshpanov2022kaznerd}, a fine-grained corpus of 112,702 annotated tokens spanning 25 entity classes. NER always runs on the toolkit's own tokens, and the tag of a multi-word token is spread over its words. Named entity spans are stored as structured objects on the document entity list with character offsets and type labels.

\subsection{Sentence Embeddings and Machine Translation}
\label{subsec:nllb}

TurkicNLP integrates NLLB-200 \cite{costa2022nllb}%
\footnote{\url{https://huggingface.co/facebook/nllb-200-distilled-600M}}
for cross-lingual sentence embeddings and machine translation for the eleven Turkic languages that it covers (Turkish, Azerbaijani, South Azerbaijani, Turkmen, Kazakh, Kyrgyz, Tatar, Bashkir, Crimean Tatar, Uzbek, and Uyghur). The \texttt{embeddings} processor mean-pools the encoder states into L2-normalized sentence and document vectors for cross-lingual similarity, retrieval, or classification. The \texttt{translate} processor translates into any of the 200 NLLB languages, specified with an ISO~639-3 or FLORES-200 code in \texttt{translate\_tgt\_lang}; the beam size is configurable via \texttt{num\_beams}. Both processors share a single distilled 600M-parameter model in memory and are installed with \texttt{pip install turkicnlp[translation]}.

%% file: sections/06-evaluation.tex
\section{Evaluation}
\label{sec:evaluation}

\begin{table*}[t]
  \centering
  \footnotesize
  \setlength{\tabcolsep}{4.5pt}
  \begin{tabular}{@{}ll l  rrr  rrr@{}}
    \toprule
    & & & \multicolumn{3}{c}{\textbf{Apertium FST}} & \multicolumn{3}{c}{\textbf{Neural (Glot500)}} \\
    \cmidrule(lr){4-6} \cmidrule(lr){7-9}
    \textbf{Language} & \textbf{ISO} & \textbf{Scr} &
    \textbf{Det\%} & \textbf{Ana\%} & \textbf{Feat\%} &
    \textbf{Det\%} & \textbf{Ana\%} & \textbf{Feat\%} \\
    \midrule
    Turkish          & tur & Latn &  \textbf{90} & 100 &  87 &  80 & 100 & \textbf{90} \\
    Azerbaijani      & aze & Latn &   50 &  53 &  42 & \textbf{75} & 100 & \textbf{79} \\
    Turkmen          & tuk & Latn &  \textbf{80} &  69 &  51 &  75 & 100 & \textbf{90} \\
    Gagauz           & gag & Latn &  \textbf{85} &  90 &  77 &  75 & 100 & \textbf{94} \\
    \cmidrule(lr){1-9}
    Kazakh           & kaz & Cyrl &  \textbf{80} & 100 &  78 &  70 & 100 & \textbf{82} \\
    Kyrgyz           & kir & Cyrl &  \textbf{80} & 100 &  81 &  60 & 100 & \textbf{92} \\
    Tatar            & tat & Cyrl &   65 &  89 &  50 & \textbf{75} & 100 & \textbf{89} \\
    Bashkir          & bak & Cyrl &   75 &  98 &  60 &  75 & 100 & \textbf{89} \\
    Crimean Tatar    & crh & Latn &  \textbf{85} &  95 &  75 &  75 & 100 & \textbf{89} \\
    Karakalpak$^\dagger$       & kaa & Latn &  \textbf{85} &  78 &  60 &  50 & 100 & \textbf{86} \\
    Nogai$^\dagger$            & nog & Cyrl &  \textbf{65} &  74 &  61 &  55 & 100 & \textbf{93} \\
    Kumyk            & kum & Cyrl &   70 &  85 &  74 &  70 & 100 & \textbf{95} \\
    Karachay-Balkar$^\dagger$  & krc & Cyrl &  \textbf{80} &  92 &  82 &  70 & 100 & \textbf{97} \\
    \cmidrule(lr){1-9}
    Uzbek            & uzb & Latn &  \textbf{85} &  98 &  83 &  55 & 100 & \textbf{87} \\
    Uyghur           & uig & Arab &  \textbf{80} & 100 & \textbf{85} &  55 & 100 &  60 \\
    \cmidrule(lr){1-9}
    Sakha            & sah & Cyrl &  \textbf{70} &  74 &  68 &  60 & 100 & \textbf{85} \\
    Altai            & alt & Cyrl &   40 &  38 &  33 & \textbf{75} & 100 & \textbf{89} \\
    Tuvan            & tyv & Cyrl &   70 &  94 &  73 & \textbf{75} & 100 & \textbf{89} \\
    Khakas           & kjh & Cyrl &   55 &  29 &  23 & \textbf{60} & 100 & \textbf{89} \\
    \cmidrule(lr){1-9}
    Chuvash          & chv & Cyrl &  \textbf{80} & 100 &  63 &  60 & 100 & \textbf{92} \\
    \midrule
    \textbf{Average} & & &  \textbf{74} &  83 &  65 &  67 &  \textbf{100} &  \textbf{88} \\
    \bottomrule
  \end{tabular}
  \caption{Morphological analysis on the 20-sentence parallel challenge set (Appendix~\ref{sec:app-challenge}). \textbf{Det\%}: sentences whose target UD feature is detected. \textbf{Ana\%}: content words with a UPOS other than \texttt{X}. \textbf{Feat\%}: content words with at least one feature. Bold: better backend per language and metric. Groups: Oghuz, Kipchak, Karluk, Siberian, Oghur. $^\dagger$Zero-shot for the neural model (no training data; proxy language embedding).}
  \label{tab:morph-eval}
\end{table*}

\begin{table*}[t]
  \centering
  \footnotesize
  \setlength{\tabcolsep}{4pt}
  \begin{tabular}{@{}ll r  rrr r  rrr@{}}
    \toprule
    & & & \multicolumn{3}{c}{\textbf{Stanza (per-language)}} & & \multicolumn{3}{c}{\textbf{Glot500 (multilingual)}} \\
    \cmidrule(lr){4-6} \cmidrule(lr){8-10}
    \textbf{Language} & \textbf{ISO} & $n$ &
    \textbf{UPOS} & \textbf{UAS} & \textbf{LAS} & &
    \textbf{UPOS} & \textbf{UAS} & \textbf{LAS} \\
    \midrule
    Turkish        & tur & 877  & 85.1 & 75.3 & 63.5 & & \textbf{86.9} & \textbf{84.5} & \textbf{70.1} \\
    Azerbaijani    & aze & 869  & 84.9 & 72.3 & 60.8 & & \textbf{87.0} & \textbf{76.2} & \textbf{64.5} \\
    Turkmen$^\ddagger$ & tuk & 825  & \textbf{86.1} & 76.8 & 67.8 & & 85.8 & \textbf{77.9} & \textbf{68.4} \\
    \cmidrule(lr){1-10}
    Kazakh         & kaz & 851  & 87.9 & 67.4 & 57.4 & & \textbf{88.8} & \textbf{71.9} & \textbf{59.9} \\
    Kyrgyz         & kir & 1166 & 71.1 & 50.2 & 29.4 & & \textbf{79.0} & \textbf{63.9} & \textbf{41.0} \\
    Bashkir$^\ddagger$ & bak & 814  & 87.0 & 72.8 & 62.5 & & \textbf{89.6} & \textbf{75.8} & \textbf{63.8} \\
    Tatar          & tat &  29  & \textbf{86.2} & 69.2 & 57.7 & & 79.3 & \textbf{77.8} & \textbf{70.4} \\
    \cmidrule(lr){1-10}
    Uyghur         & uig & 758  & 76.9 & 77.1 & 62.3 & & \textbf{79.7} & \textbf{81.5} & \textbf{66.7} \\
    Uzbek          & uzb & 905  & 71.8 & 67.2 & 46.3 & & \textbf{87.6} & \textbf{81.5} & \textbf{68.3} \\
    \cmidrule(lr){1-10}
    \rowcolor{codebg}
    Karakalpak$^\dagger$     & kaa &  235 & --- & --- & --- & & 77.5 & 61.7 & 45.8 \\
    \rowcolor{codebg}
    Kumyk$^\dagger$          & kum & 144  & --- & --- & --- & & 78.5 & 60.4 & 45.5 \\
    \rowcolor{codebg}
    Sakha$^\dagger$          & sah &  16  & --- & --- & --- & & 93.8 & 60.0 & 46.7 \\
    \midrule
    \textbf{Avg (9 shared)} & & & 81.9 & 69.8 & 56.4 & & \textbf{84.9} & \textbf{76.8} & \textbf{63.7} \\
    \bottomrule
  \end{tabular}
  \caption{POS tagging and dependency parsing. Both backends receive the same automatic tokenization; $n$: test words whose boundaries match the gold tokenization, over which UPOS is computed; UAS and LAS are computed over those of these words whose gold and predicted heads are also matched. Bold: better backend per language and metric. $^\dagger$Zero-shot (shaded): no training data and no Stanza model; Glot500 uses a proxy language embedding (Karakalpak$\to$Uzbek, Kumyk$\to$Tatar, Sakha$\to$Bashkir). $^\ddagger$Silver test set (translated and automatically annotated). Averages: nine shared languages.}
  \label{tab:parsing-eval}
\end{table*}

In this section, we evaluate the components for which TurkicNLP offers competing backends, i.e., finite-state versus neural morphological analysis and per-language versus multilingual parsing. We then evaluate the script layer and examine whether a common script reduces the tokenization gap described in \S\ref{subsec:related-tokenizer}. The evaluation code, data, and outputs are released with the toolkit.

\subsection{Morphological Analysis Across 20 Languages}
\label{subsec:eval-morph}

Since gold morphological annotation is missing for most Turkic languages, we built a parallel challenge set of 20 sentences translated into 20 languages (Appendix~\ref{sec:app-challenge}), each targeting one phenomenon (five cases; past, progressive, future, negation, question, imperative, conditional, evidential, passive; plural, possessive, pronoun, reciprocal, numeral, converb). Table~\ref{tab:morph-eval} reports \textbf{Det\%}, the share of sentences in which the target UD feature (e.g., \texttt{Case=Acc}) is assigned to some token, and two coverage measures: \textbf{Ana\%}, content words with a UPOS other than \texttt{X} (100\% by construction for the neural classifier), and \textbf{Feat\%}, content words with at least one feature.

On average, the transducers detect the target phenomenon more often than the neural analyzer (74\% vs.\ 67\%), whereas the neural analyzer labels every content word and assigns features to many more of them (88\% vs.\ 65\%). The neural model is ahead where the transducers are prototypes with small lexicons, most clearly for Altai (40\% vs.\ 75\%) and Azerbaijani (50\% vs.\ 75\%), and it gives usable analyses even zero-shot (Karachay-Balkar 70\%, Nogai 55\%). Apertium stays ahead for the languages with mature transducers (Turkish, Kazakh, Kyrgyz, Uzbek, and Uyghur), for Karakalpak, which the neural model has not seen, and for Chuvash, whose divergent Oghur morphology seems to limit transfer from the other branches. The two backends are therefore complementary, and the morpheme segmenter (\S\ref{subsec:morph-seg}) exploits this. On a segmentation test set of 483 words in 13 word categories and 20 languages (Appendix~\ref{sec:app-morphseg}), combining both analyzers raises the share of exactly segmented words from 68\% (neural only) and 79\% (Apertium only) to 89\%. The neural analyzer alone often misplaces the stem boundary in verbs and almost never splits derivational suffixes, while the transducers leave words outside their lexicons unanalyzed; the hybrid is best or tied in every word category except unsegmentable words (Table~\ref{tab:morphseg-cat}).

\subsection{POS Tagging and Dependency Parsing}
\label{subsec:eval-parsing}

We compare the per-language Stanza models with the multilingual Glot500 parser on 12 languages. Both backends receive the same automatic tokenization (token F1 between 0.95 and 1.00 on the languages they share), and we score them on the words whose boundaries match the gold segmentation. Gold test sets come from the TueCL parallel treebanks of the UD-Turkic initiative: the published Azerbaijani, Kyrgyz, Turkish, and Uzbek treebanks \cite{eslami-etal-2025-parallel} and in-progress files for Kazakh, Uyghur, Tatar, Karakalpak (35 translated sentences), Kumyk, and Sakha.\footnote{\url{https://github.com/ud-turkic/parallel}} For Turkmen and Bashkir, which lack gold data, we use silver sets obtained by translating and automatically annotating the Turkish test set (\S\ref{subsec:pos-dep}), so these results are only indicative.

Table~\ref{tab:parsing-eval} shows that sharing parameters across the family is beneficial: the multilingual parser outperforms the per-language models on average in UPOS (84.9 vs.\ 81.9), UAS (76.8 vs.\ 69.8), and LAS (63.7 vs.\ 56.4), with a higher LAS for all nine shared languages. Gains are largest for small or noisy treebanks, i.e., Uzbek (+22.0 LAS), Tatar (+12.7), and Kyrgyz (+11.6), but even Turkish, which has by far the most training data, gains 6.6 LAS. Without any target-language data, the parser reaches 45.5--46.7 LAS for Karakalpak, Kumyk, and Sakha, which is close to the supervised Stanza model for Uzbek and well above the one for Kyrgyz, although the zero-shot test sets are small. Stanza retains a small UPOS advantage for Turkmen (86.1 vs.\ 85.8) and Tatar (86.2 vs.\ 79.3), although the Tatar test set contains only 29 words.

\subsection{Script Layer}
\label{subsec:eval-scripts}

\paragraph{Transliteration.} Lacking gold parallel-script text, we measure \emph{round-trip consistency}: native-script text from the challenge set and TueCL is converted to the other script and back, and we count the words recovered exactly (Table~\ref{tab:translit-eval}, Appendix~\ref{sec:app-script-results}). Conversion is fully reversible for all languages and pivots except Crimean Tatar (97.6\%), whose Cyrillic orthography writes word-initial \textit{ö} and \textit{ü} as \textit{о} and \textit{у} (\textit{ötmek} returns as \textit{otmek}), and at least 99.7\% of the letters in the intermediate output belong to the target script. The round trips also served as regression tests, which revealed errors in the position-dependent rules for Kazakh, Uzbek, Turkmen, and Tatar that we subsequently fixed. Our Uzbek conversions agree with the independent UzTransliterator\footnote{\url{https://pypi.org/project/UzTransliterator/}} on 99.4\% of the words for Cyrillic to Latin and 98.4\% for Latin to Cyrillic. The remaining differences are Russian loanwords that would require exception lists (e.g., \textit{film} $\to$ \textit{фильм}).

\paragraph{Script detection.} On the 400 native challenge sentences and their 580 conversions into all supported scripts, \texttt{detect\_script} is correct for every sentence and word (3,147 words). It also recovers both scripts and the exact boundary in 500 synthetic mixed-script texts, is not affected by one or two Latin homoglyphs in Cyrillic text, and recognizes all 73 letters of the Old Turkic runic block.

\paragraph{Language identification.} GlotLID identifies the larger languages almost perfectly (Turkish, Kazakh, Kyrgyz, Bashkir, Uzbek, and Uyghur; Table~\ref{tab:langid-eval}, Appendix~\ref{sec:app-script-results}). Overall, it reaches 85.7\% on the natural TueCL sentences but only 69.2\% on the short challenge sentences. The errors show where the family is hardest to separate, as smaller languages are absorbed by a close neighbor in the same script: Gagauz is labeled Turkish in 18 of 20 cases, Crimean Tatar is confused with Turkish, Khakas with Kazakh, Nogai, Altai, and Tuvan with Kyrgyz, and Kumyk with Karachay-Balkar. Karakalpak in the 2016 Latin orthography is recognized in none of the challenge sentences and in only 34.3\% of the TueCL sentences. Restricting the label space to Turkic varieties never hurts and yields small gains (+2.0 and +0.6 points). On script-converted text, however, accuracy drops to 48.1\%, which suggests that language identification should be applied before transliteration.

\subsection{Does a Common Script Narrow the Tokenization Gap?}
\label{subsec:eval-cts}

We convert the parallel sentences of \S\ref{subsec:related-tokenizer} to CTS and measure fertility again (Table~\ref{tab:cts-fertility}, Appendix~\ref{sec:app-script-results}). For the 13 languages written in Cyrillic or Arabic script, CTS reduces the average fertility by 22--41\%: from 3.08 to 2.16 for Qwen3.5, 3.64 to 2.27 for LLaMA~3, 4.17 to 2.45 for OLMo~3, and 2.72 to 2.14 for GPT-5.2. The largest reductions are observed for Uyghur (6.85 to 2.64 under OLMo~3), Bashkir, and Sakha, while the Latin-script languages remain almost unchanged. The difference between the two groups thus shrinks from 0.62--1.70 to at most 0.03 tokens per word, which indicates that the gap is almost entirely a script effect. CTS is roughly as compact as the national Latin orthographies; under Qwen3.5, for instance, it is more compact for Kazakh (2.28 vs.\ 2.48 in the 2021 Latin alphabet) and less compact for Uzbek (2.30 vs.\ 2.12). A shared script also makes cognates visible: for language pairs written in different scripts, the character-trigram overlap between their parallel sentences rises from 0 to 0.30 on average (Turkish--Kumyk 0.42), and word-type overlap more than doubles.

\paragraph{Downstream effects.} Lower fertility alone does not imply better model quality, since LLMs have seen little CTS text during training. We leave the evaluation of downstream effects, e.g., of continued pretraining or tokenizer adaptation on CTS text, to future work.

\subsection{Named Entity Recognition}
\label{subsec:eval-ner}
A systematic evaluation of NER (entity-level F1 for Turkish and Kazakh) is planned for a future version of this paper.

%% file: sections/10-conclusion.tex
\section{Conclusion}
\label{sec:conclusion}

We presented TurkicNLP, an open-source toolkit that brings fragmented resources for Turkic languages, i.e., finite-state transducers, parsers trained on UD treebanks, multilingual encoders, and translation models, together in a single script-aware pipeline for 24 languages. The toolkit offers one interface for every language and backend, handles the orthographic diversity of the family automatically, and keeps its core lightweight through catalog-driven model management. Our evaluation supports its cross-lingual design. The multilingual parser achieves a higher LAS than the per-language parsers for all nine languages covered by both, the multilingual morphological analyzer extends analysis to languages that only have prototype transducers, and both models transfer zero-shot to languages without training data. The script layer converts text reliably, and conversion to the Common Turkic Alphabet removes most of the script-driven tokenization gap of LLM tokenizers. We hope that TurkicNLP lowers the barrier to working with the more than 200 million speakers of Turkic languages, in particular with the many languages that so far have little or no digital NLP infrastructure.

%% file: sections/08-future.tex
\section{Future Directions}
\label{sec:future}

Our next steps are gold-standard evaluations of morphological features, lemmas, NER, and transliteration, which require annotated data that does not yet exist for most of the languages; morphology-aware tokenizers trained on corpora segmented with the hybrid segmenter; evaluating whether text in the Common Turkic Alphabet also improves the downstream performance of language models; and training infrastructure that lets researchers contribute models for further languages.

%% file: sections/09-limitations.tex
\section*{Limitations}
\label{sec:limitations}

The first limitation concerns coverage, which differs considerably between languages. Only Turkish and Kazakh have the full pipeline including NER, while several languages rely on prototype transducers, on training with paradigms only, or on zero-shot transfer (Table~\ref{tab:coverage-matrix}).

Another limitation concerns the evaluation data. The morphological evaluation uses a small parallel challenge set with a lenient detection criterion and coverage measures instead of token-level precision and recall against gold annotations, which do not exist for most of the languages. It therefore indicates the relative behavior of the backends rather than their absolute accuracy. Several parsing test sets are very small (Tatar, Sakha, Kumyk, Karakalpak), two are silver (Turkmen, Bashkir), and some of the custom Stanza models were trained on silver data, so these results carry substantial uncertainty and possibly annotation bias. The morpheme segmentation test set is also small, its gold segmentations for nine languages are drafts that still await verification by native speakers, and it was used during development, so its scores are likely optimistic. The language identification, CTS, and tokenizer experiments are based on small parallel sentence sets and are meant as diagnostics, and lower fertility under CTS does not by itself show better downstream performance, since LLMs have seen little CTS text.

We did not evaluate NER in this paper, and we evaluated transliteration through round-trip consistency and, for Uzbek, through agreement with an independent converter rather than against gold parallel-script text. Some errors are inherent to the orthographies: word-initial \textit{ö} and \textit{ü} in Crimean Tatar cannot be recovered from Cyrillic, and loanwords with lexical exceptions (e.g., Russian soft signs in Uzbek Cyrillic) are not handled (\S\ref{subsec:eval-scripts}). Transliteration can also be ambiguous for proper nouns and code-mixed text, and the conversion of Ottoman Turkish (from Latin to Perso-Arabic only) does not yet map transcription diacritics such as \textit{â} or \textit{ḥ}. The disambiguation of transducer readings is heuristic. Finally, the NLLB-based embeddings and translation are limited to eleven languages and are distributed under a non-commercial license.

\section*{Ethical Considerations}
The choice of script is politically and culturally sensitive in several Turkic-speaking regions. TurkicNLP implements official and community standards side by side and makes every conversion optional. It does not promote any orthography, and we offer the Common Turkic Alphabet only as a technical layer for interoperability. Components with lower quality tiers may produce incorrect analyses for under-resourced languages, which is why the quality tiers are listed in the model catalog. Third-party data and models keep their original licenses (GPL-3.0 for Apertium, CC-BY-NC-4.0 for NLLB) and are downloaded separately from the Apache-2.0 library.

%% file: sections/appendix-d-resources.tex
\definecolor{hdrTrn}{RGB}{210,232,250}  
\definecolor{hdrBnc}{RGB}{232,218,252}  
\definecolor{hdrNlp}{RGB}{210,240,222}  
\definecolor{hdrTkl}{RGB}{252,232,212}  
\definecolor{branchbg}{RGB}{232,232,232}

\onecolumn
\begin{landscape}
\setlength{\tabcolsep}{2.5pt}
\renewcommand{\arraystretch}{1.15}

\begin{longtable}{%
  >{\raggedright\arraybackslash\tiny}p{2.0cm}|%
  >{\raggedright\arraybackslash\tiny}p{1.35cm}%
  >{\raggedright\arraybackslash\tiny}p{1.65cm}%
  >{\raggedright\arraybackslash\tiny}p{1.5cm}%
  >{\raggedright\arraybackslash\tiny}p{1.5cm}|%
  >{\raggedright\arraybackslash\tiny}p{1.75cm}%
  >{\raggedright\arraybackslash\tiny}p{1.35cm}|%
  >{\raggedright\arraybackslash\tiny}p{1.5cm}%
  >{\raggedright\arraybackslash\tiny}p{1.6cm}%
  >{\raggedright\arraybackslash\tiny}p{1.5cm}%
  >{\raggedright\arraybackslash\tiny}p{1.4cm}%
  >{\raggedright\arraybackslash\tiny}p{1.4cm}%
  >{\raggedright\arraybackslash\tiny}p{1.4cm}|%
  >{\raggedright\arraybackslash\tiny}p{1.95cm}}

\caption{Comprehensive overview of NLP resources for 24 Turkic languages.
Each cell lists available tools/datasets as references; \textbf{---} indicates
no known public resource. Branches: Oghuz, Kipchak, Karluk, Siberian, Oghur, Arghu, Historical.
Apertium quality tiers: \emph{Prod.}~=~Production, \emph{Stb.}~=~Stable, \emph{Beta},
\emph{Proto.}~=~Prototype.}
\label{tab:resources-full}\\

\toprule
\multirow{2}{*}{\normalsize\textbf{Language}} &
\multicolumn{4}{c|}{\cellcolor{hdrTrn}\textbf{Datasets --- Training}} &
\multicolumn{2}{c|}{\cellcolor{hdrBnc}\textbf{Datasets --- Benchmarks}} &
\multicolumn{6}{c|}{\cellcolor{hdrNlp}\textbf{NLP Components}} &
\multicolumn{1}{c}{\cellcolor{hdrTkl}\textbf{Toolkits}} \\
\cmidrule(lr){2-5}\cmidrule(lr){6-7}\cmidrule(lr){8-13}\cmidrule(l){14-14}
& \cellcolor{hdrTrn}\textbf{UD Tree-banks}
& \cellcolor{hdrTrn}\textbf{NER Corpus}
& \cellcolor{hdrTrn}\textbf{Parallel/ MT Corpus}
& \cellcolor{hdrTrn}\textbf{Speech Corpus}
& \cellcolor{hdrBnc}\textbf{NLU Bench.}
& \cellcolor{hdrBnc}\textbf{QA}
& \cellcolor{hdrNlp}\textbf{Morphology}
& \cellcolor{hdrNlp}\textbf{POS + Dep.}
& \cellcolor{hdrNlp}\textbf{NER System}
& \cellcolor{hdrNlp}\textbf{ASR}
& \cellcolor{hdrNlp}\textbf{TTS}
& \cellcolor{hdrNlp}\textbf{MT}
& \cellcolor{hdrTkl}\textbf{Toolkit} \\
\midrule
\endfirsthead

\multicolumn{14}{l}{\tiny\textit{\dots continued from previous page}}\\[2pt]
\toprule
\multirow{2}{*}{\normalsize\textbf{Language}} &
\multicolumn{4}{c|}{\cellcolor{hdrTrn}\textbf{Datasets --- Training}} &
\multicolumn{2}{c|}{\cellcolor{hdrBnc}\textbf{Datasets --- Benchmarks}} &
\multicolumn{6}{c|}{\cellcolor{hdrNlp}\textbf{NLP Components}} &
\multicolumn{1}{c}{\cellcolor{hdrTkl}\textbf{Toolkits}} \\
\cmidrule(lr){2-5}\cmidrule(lr){6-7}\cmidrule(lr){8-13}\cmidrule(l){14-14}
& \cellcolor{hdrTrn}\textbf{UD Tree-banks}
& \cellcolor{hdrTrn}\textbf{NER Corpus}
& \cellcolor{hdrTrn}\textbf{Parallel/ MT Corpus}
& \cellcolor{hdrTrn}\textbf{Speech Corpus}
& \cellcolor{hdrBnc}\textbf{NLU Bench.}
& \cellcolor{hdrBnc}\textbf{QA}
& \cellcolor{hdrNlp}\textbf{Morphology}
& \cellcolor{hdrNlp}\textbf{POS + Dep.}
& \cellcolor{hdrNlp}\textbf{NER System}
& \cellcolor{hdrNlp}\textbf{ASR}
& \cellcolor{hdrNlp}\textbf{TTS}
& \cellcolor{hdrNlp}\textbf{MT}
& \cellcolor{hdrTkl}\textbf{Toolkit} \\
\midrule
\endhead

\midrule
\multicolumn{14}{r}{\tiny\textit{Continued on next page\dots}}\\
\endfoot

\bottomrule
\endlastfoot

\multicolumn{14}{l}{\cellcolor{branchbg}\textit{Oghuz Branch}} \\
\midrule

Turkish (tur)
&
\citep{sulubacak-etal-2016-universal};\newline \citep{turk-etal-2019-turkish};\newline \citep{ddi_itu}
&
MilliyetNER \citep{tur2003milliyetner};\newline TWNERTC \citep{sahin2017twnertc}&
NLLB \citep{costa2022nllb};\newline OPUS \citep{til_github};\newline Leipzig \citep{goldhahn2012leipzig}
&
CommonVoice \citep{ardila-etal-2020-common};\newline TSC \citep{tsc2024turkish};\newline FLEURS \citep{conneau2022fleurs}
&
Mukayese \citep{safaya-etal-2022-mukayese};\newline TR-MMLU \citep{bayram2025trmmlu};\newline TurkishMMLU \citep{yuksel2024turkishmmlu};\newline Cetvel \citep{abrek2025cetvel};\newline TurkBench \citep{toraman2026turkbench};\newline TUMLU \citep{isbarov2025tumlu}
&
---
&
Apertium Prod. \citep{forcada2011apertium};\newline Zemberek \citep{akin2007zemberek}
&
Stanza \citep{qi2020stanza};\newline TurkishDelightNLP \citep{alecakir-etal-2022-turkishdelightnlp};\newline ITU pipeline \citep{ddi_itu}
&
TurkishDelightNLP \citep{alecakir-etal-2022-turkishdelightnlp};\newline Stanza \citep{qi2020stanza}
&
CommonVoice \citep{ardila-etal-2020-common};\newline S{\"o}yle \citep{issai2023soyle};\newline TurkicASR \citep{mussakhojayeva2023multilingual}
&
TurkicTTS \citep{yeshpanov2023turkictts}
&
NLLB \citep{costa2022nllb};\newline Apertium \citep{forcada2011apertium}
&
Zemberek \citep{akin2007zemberek};\newline TurkishDelightNLP \citep{alecakir-etal-2022-turkishdelightnlp};\newline ITU NLP \citep{itunlp};\newline TurkNLP Suite \citep{turkish_nlp_suite}
\\[2pt]

Azerbaijani (aze)
&
\citep{eslami-etal-2025-parallel}
&
azWikiNER \citep{ibiyev2021azwikiner};\newline AzNER \citep{localdoc2024azner}&
NLLB \citep{costa2022nllb};\newline TIL \citep{til_github};\newline Leipzig \citep{goldhahn2012leipzig}
&
Multilingual \citep{mussakhojayeva2023multilingual};\newline AzASR \citep{localdoc2025azasr};\newline FLEURS \citep{conneau2022fleurs}
&
TUMLU \citep{isbarov2025tumlu};\newline Karde\c{s}-NLU \citep{senel2024kardes}
&
---
&
Apertium Stb. \citep{forcada2011apertium}
&
UDPipe \citep{straka2016udpipe}
&
XLM-R NER \citep{samadov2024azner}
&
TurkicASR \citep{mussakhojayeva2023multilingual}
&
TurkicTTS \citep{yeshpanov2023turkictts}
&
NLLB \citep{costa2022nllb}
&
---
\\[2pt]

Turkmen (tuk)
&
---
&
---
&
NLLB \citep{costa2022nllb};\newline TIL \citep{til_github};\newline Leipzig \citep{goldhahn2012leipzig}
&
TurkmenSpeech \citep{mamedov2024turkmenspeech}
&
---
&
---
&
Apertium Beta \citep{forcada2011apertium}
&
---
&
---
&
S{\"o}yle \citep{issai2023soyle}
&
TurkicTTS \citep{yeshpanov2023turkictts}
&
NLLB \citep{costa2022nllb}
&
---
\\[2pt]

Gagauz (gag)
&
---
&
---
&
NLLB \citep{costa2022nllb}
&
---
&
---
&
---
&
Apertium Proto. \citep{forcada2011apertium}
&
---
&
---
&
---
&
---
&
NLLB \citep{costa2022nllb}
&
---
\\[2pt]

Iranian Azerbaijani (azb)
&
---
&
---
&
NLLB \citep{costa2022nllb}
&
---
&
---
&
---
&
---
&
---
&
---
&
---
&
---
&
NLLB \citep{costa2022nllb}
&
---
\\[4pt]

\multicolumn{14}{l}{\cellcolor{branchbg}\textit{Kipchak Branch}} \\
\midrule

Kazakh (kaz)
&
\citep{tyers-washington-2015-kazakh}
&
KazNERD \citep{yeshpanov2022kaznerd}&
NLLB \citep{costa2022nllb};\newline TIL \citep{til_github};\newline Leipzig \citep{goldhahn2012leipzig}
&
KazakhTTS2 \citep{mussakhojayeva-etal-2022-kazakhtts2};\newline KSC2 \citep{mussakhojayeva2023turkksc};\newline Multilingual \citep{mussakhojayeva2023multilingual};\newline FLEURS \citep{conneau2022fleurs};\newline CommonVoice \citep{ardila-etal-2020-common}
&
TUMLU \citep{isbarov2025tumlu};\newline Karde\c{s}-NLU \citep{senel2024kardes}
&
KazQAD \citep{yeshpanov2024kazqad}
&
Apertium Prod. \citep{forcada2011apertium}
&
Stanza \citep{qi2020stanza}
&
KazNERD \citep{yeshpanov2022kaznerd};\newline Stanza \citep{qi2020stanza}
&
TurkicASR \citep{mussakhojayeva2023multilingual}
&
KazakhTTS2 \citep{mussakhojayeva-etal-2022-kazakhtts2};\newline TurkicTTS \citep{yeshpanov2023turkictts}
&
NLLB \citep{costa2022nllb};\newline TIL \citep{til_github};\newline Apertium \citep{forcada2011apertium}
&
KazNLP \citep{kaznlp_github,makhambetov-etal-2013-assembling}
\\[2pt]

Kyrgyz (kir)
&
\citep{eslami-etal-2025-parallel}
&
KyrgyzNER \citep{turatali2025humanannotatednerdatasetkyrgyz}&
NLLB \citep{costa2022nllb};\newline TIL \citep{til_github};\newline Leipzig \citep{goldhahn2012leipzig}
&
Multilingual \citep{mussakhojayeva2023multilingual};\newline CommonVoice \citep{ardila-etal-2020-common};\newline FLEURS \citep{conneau2022fleurs}
&
TUMLU \citep{isbarov2025tumlu};\newline Karde\c{s}-NLU \citep{senel2024kardes};\newline \citep{11206960}
&
---
&
Apertium Stb. \citep{forcada2011apertium}
&
Stanza \citep{qi2020stanza}
&
Stanza \citep{qi2020stanza}
&
TurkicASR \citep{mussakhojayeva2023multilingual};\newline AkylAI-STT \citep{cramerproject2024akylai}
&
TurkicTTS \citep{yeshpanov2023turkictts};\newline AkylAI-TTS \citep{cramerproject2024akylai}
&
NLLB \citep{costa2022nllb};\newline TIL \citep{til_github}
&
---
\\[2pt]

Tatar (tat)
&
\citep{taguchi-etal-2022-universal}
&
---
&
NLLB \citep{costa2022nllb};\newline TIL \citep{til_github};\newline Leipzig \citep{goldhahn2012leipzig}
&
TatSC \citep{tatsc2023tatar};\newline TatarTTS \citep{orel2024tatartts};\newline CommonVoice \citep{ardila-etal-2020-common};\newline Multilingual \citep{mussakhojayeva2023multilingual}
&
TUMLU \citep{isbarov2025tumlu}
&
---
&
Apertium Prod. \citep{forcada2011apertium}
&
UDPipe \citep{straka2016udpipe}
&
---
&
S{\"o}yle \citep{issai2023soyle};\newline TurkicASR \citep{mussakhojayeva2023multilingual}
&
TurkicTTS \citep{yeshpanov2023turkictts}
&
NLLB \citep{costa2022nllb};\newline Apertium \citep{forcada2011apertium}
&
---
\\[2pt]

Bashkir (bak)
&
---
&
---
&
NLLB \citep{costa2022nllb};\newline Leipzig \citep{goldhahn2012leipzig}
&
CommonVoice \citep{ardila-etal-2020-common};\newline Multilingual \citep{mussakhojayeva2023multilingual}
&
---
&
---
&
Apertium Beta \citep{forcada2011apertium}
&
---
&
---
&
TurkicASR \citep{mussakhojayeva2023multilingual}
&
TurkicTTS \citep{yeshpanov2023turkictts}
&
NLLB \citep{costa2022nllb}
&
---
\\[2pt]

Crimean Tatar (crh)
&
---
&
---
&
NLLB \citep{costa2022nllb};\newline Apertium \citep{forcada2011apertium}
&
---
&
TUMLU \citep{isbarov2025tumlu}
&
---
&
Apertium Beta \citep{forcada2011apertium}
&
---
&
---
&
---
&
---
&
NLLB \citep{costa2022nllb};\newline Apertium \citep{forcada2011apertium}
&
---
\\[2pt]

Karakalpak (kaa)
&
---
&
---
&
NLLB \citep{costa2022nllb};\newline TIL \citep{til_github}
&
---
&
TUMLU \citep{isbarov2025tumlu}
&
---
&
Apertium Proto. \citep{forcada2011apertium}
&
---
&
---
&
---
&
---
&
NLLB \citep{costa2022nllb}
&
---
\\[2pt]

Nogai (nog)
&
---
&
---
&
NLLB \citep{costa2022nllb}
&
---
&
---
&
---
&
Apertium Proto. \citep{forcada2011apertium}
&
---
&
---
&
---
&
---
&
NLLB \citep{costa2022nllb}
&
---
\\[2pt]

Kumyk (kum)
&
---
&
---
&
NLLB \citep{costa2022nllb}
&
---
&
---
&
---
&
---
&
---
&
---
&
---
&
---
&
NLLB \citep{costa2022nllb}
&
---
\\[2pt]

Karachay-Balkar (krc)
&
---
&
---
&
NLLB \citep{costa2022nllb};\newline Leipzig \citep{goldhahn2012leipzig}
&
---
&
---
&
---
&
---
&
---
&
---
&
---
&
---
&
NLLB \citep{costa2022nllb}
&
---
\\[4pt]

\multicolumn{14}{l}{\cellcolor{branchbg}\textit{Karluk Branch}} \\
\midrule

Uzbek (uzb)
&
\citep{akhundjanova-talamo-2025-uzbek}
&
UZNER \citep{yusufu2023uzner}&
NLLB \citep{costa2022nllb};\newline TIL \citep{til_github};\newline Leipzig \citep{goldhahn2012leipzig}
&
USC \citep{usc2021};\newline Multilingual \citep{mussakhojayeva2023multilingual};\newline FLEURS \citep{conneau2022fleurs}
&
TUMLU \citep{isbarov2025tumlu};\newline Karde\c{s}-NLU \citep{senel2024kardes}
&
---
&
Apertium Stb. \citep{forcada2011apertium}
&
UDPipe \citep{straka2016udpipe}
&
---
&
TurkicASR \citep{mussakhojayeva2023multilingual}
&
TurkicTTS \citep{yeshpanov2023turkictts}
&
NLLB \citep{costa2022nllb};\newline TIL \citep{til_github}
&
---
\\[2pt]

Uyghur (uig)
&
\citep{eli-etal-2016-universal}
&
---
&
NLLB \citep{costa2022nllb};\newline TIL \citep{til_github};\newline Leipzig \citep{goldhahn2012leipzig}
&
CommonVoice \citep{ardila-etal-2020-common};\newline Multilingual \citep{mussakhojayeva2023multilingual}
&
TUMLU \citep{isbarov2025tumlu};\newline Karde\c{s}-NLU \citep{senel2024kardes}
&
---
&
Apertium Beta \citep{forcada2011apertium}
&
Stanza \citep{qi2020stanza}
&
---
&
TurkicASR \citep{mussakhojayeva2023multilingual}
&
TurkicTTS \citep{yeshpanov2023turkictts}
&
NLLB \citep{costa2022nllb}
&
---
\\[4pt]

\multicolumn{14}{l}{\cellcolor{branchbg}\textit{Siberian Branch}} \\
\midrule

Sakha (sah)
&
\citep{merzhevich-ferraz-gerardi-2022-introducing}
&
---
&
NLLB \citep{costa2022nllb};\newline Leipzig \citep{goldhahn2012leipzig}
&
CommonVoice \citep{ardila-etal-2020-common};\newline Multilingual \citep{mussakhojayeva2023multilingual}
&
---
&
---
&
Apertium Beta \citep{forcada2011apertium}
&
---
&
---
&
TurkicASR \citep{mussakhojayeva2023multilingual}
&
TurkicTTS \citep{yeshpanov2023turkictts}
&
NLLB \citep{costa2022nllb}
&
---
\\[2pt]

Altai (alt)
&
---
&
---
&
NLLB \citep{costa2022nllb}
&
---
&
---
&
---
&
Apertium Proto. \citep{forcada2011apertium}
&
---
&
---
&
---
&
---
&
NLLB \citep{costa2022nllb}
&
---
\\[2pt]

Tuvan (tyv)
&
---
&
---
&
NLLB \citep{costa2022nllb}
&
---
&
---
&
---
&
Apertium Proto. \citep{forcada2011apertium}
&
---
&
---
&
---
&
---
&
NLLB \citep{costa2022nllb}
&
---
\\[2pt]

Khakas (kjh)
&
---
&
---
&
NLLB \citep{costa2022nllb}
&
---
&
---
&
---
&
---
&
---
&
---
&
---
&
---
&
NLLB \citep{costa2022nllb}
&
---
\\[4pt]

\multicolumn{14}{l}{\cellcolor{branchbg}\textit{Oghur Branch}} \\
\midrule

Chuvash (chv)
&
---
&
---
&
NLLB \citep{costa2022nllb};\newline Leipzig \citep{goldhahn2012leipzig}
&
CommonVoice \citep{ardila-etal-2020-common}
&
---
&
---
&
Apertium Proto. \citep{forcada2011apertium}
&
---
&
---
&
TurkicASR \citep{mussakhojayeva2023multilingual}
&
---
&
NLLB \citep{costa2022nllb}
&
---
\\[4pt]

\multicolumn{14}{l}{\cellcolor{branchbg}\textit{Arghu Branch}} \\
\midrule

Khalaj (klj)
&
---
&
---
&
NLLB \citep{costa2022nllb}
&
---
&
---
&
---
&
---
&
---
&
---
&
---
&
---
&
NLLB \citep{costa2022nllb}
&
---
\\[4pt]

\multicolumn{14}{l}{\cellcolor{branchbg}\textit{Historical Languages}} \\
\midrule

Ottoman Turkish (ota)
&
\citep{ozates-etal-2024-dependency}
&
---
&
NLLB \citep{costa2022nllb}
&
---
&
---
&
---
&
Apertium Proto. \citep{forcada2011apertium}
&
Stanza \citep{qi2020stanza}
&
---
&
---
&
---
&
NLLB \citep{costa2022nllb}
&
---
\\[2pt]

Old Turkish (otk)
&
\citep{derin-harada-2021-universal}
&
---
&
---
&
---
&
---
&
---
&
---
&
---
&
---
&
---
&
---
&
---
&
---
\\

\end{longtable}
\end{landscape}
\twocolumn

%% file: sections/llm-models-table.tex
\onecolumn
\begin{landscape}
\setlength{\tabcolsep}{2.5pt}
\renewcommand{\arraystretch}{1.15}

\definecolor{hdrEnc}{RGB}{210,232,250}   
\definecolor{hdrGen}{RGB}{252,232,212}   
\definecolor{hdrMul}{RGB}{210,240,222}   

\begin{longtable}{%
  >{\raggedright\arraybackslash\tiny}p{2.1cm}|%
  >{\raggedright\arraybackslash\tiny}p{5.8cm}|%
  >{\raggedright\arraybackslash\tiny}p{5.8cm}|%
  >{\raggedright\arraybackslash\tiny}p{5.0cm}}

\caption{Pretrained neural language models for Turkic languages.
Encoder: masked LM (BERT/RoBERTa/ELECTRA-style);
Generative: causal/autoregressive LM (GPT/LLaMA-style) and encoder-decoder (T5/BART-style);
Multilingual: large multilingual models with confirmed Turkic-language pretraining data.
\textbf{---} indicates no publicly available model.}
\label{tab:llm-models}\\

\toprule
\multirow{2}{*}{\normalsize\textbf{Language}} &
\multicolumn{1}{c|}{\cellcolor{hdrEnc}\textbf{Encoder-only (MLM)}} &
\multicolumn{1}{c|}{\cellcolor{hdrGen}\textbf{Generative / Seq2Seq}} &
\multicolumn{1}{c}{\cellcolor{hdrMul}\textbf{Multilingual models}} \\
&
\cellcolor{hdrEnc}\small BERT / RoBERTa / ELECTRA &
\cellcolor{hdrGen}\small GPT / LLaMA / T5 / BART &
\cellcolor{hdrMul}\small (coverage confirmed) \\
\midrule
\endfirsthead

\multicolumn{4}{l}{\tiny\textit{\dots continued from previous page}}\\[2pt]
\toprule
\multirow{2}{*}{\normalsize\textbf{Language}} &
\multicolumn{1}{c|}{\cellcolor{hdrEnc}\textbf{Encoder-only (MLM)}} &
\multicolumn{1}{c|}{\cellcolor{hdrGen}\textbf{Generative / Seq2Seq}} &
\multicolumn{1}{c}{\cellcolor{hdrMul}\textbf{Multilingual models}} \\
&
\cellcolor{hdrEnc}\small BERT / RoBERTa / ELECTRA &
\cellcolor{hdrGen}\small GPT / LLaMA / T5 / BART &
\cellcolor{hdrMul}\small (coverage confirmed) \\
\midrule
\endhead

\midrule
\multicolumn{4}{r}{\tiny\textit{Continued on next page\dots}}\\
\endfoot

\bottomrule
\endlastfoot

\multicolumn{4}{l}{\cellcolor{branchbg}\textit{Oghuz Branch}} \\
\midrule

Turkish (tur)
&
BERTurk (base/cased/uncased/128k; DistilBERT; ELECTRA; ConvBERT) \citep{schweter2020berturk}
&
TURNA (1.1B enc-dec) \citep{uludogan-etal-2024-turna};\newline Kumru-2B \citep{vngrs2024kumru};\newline Trendyol-LLM-8B \citep{trendyol2024llm}
&
mBERT \citep{devlin2019bert};\newline XLM-R \citep{conneau2020unsupervised};\newline Glot500 \citep{imanigooghari-etal-2023-glot500};\newline mGPT \citep{shliazhko-etal-2024-mgpt};\newline NLLB \citep{costa2022nllb}
\\[2pt]

Azerbaijani (aze)
&
aLLMA (SMALL/BASE/LARGE) \citep{hajili2024allma}
&
Az-Mistral (7B) \citep{hajili2024allma}
&
mBERT \citep{devlin2019bert};\newline XLM-R \citep{conneau2020unsupervised};\newline Glot500 \citep{imanigooghari-etal-2023-glot500};\newline mGPT \citep{shliazhko-etal-2024-mgpt};\newline NLLB \citep{costa2022nllb}
\\[2pt]

Turkmen (tuk)
&
---
&
---
&
Glot500 \citep{imanigooghari-etal-2023-glot500};\newline mGPT \citep{shliazhko-etal-2024-mgpt};\newline NLLB \citep{costa2022nllb}
\\[2pt]

Gagauz (gag)
&
---
&
---
&
Glot500 \citep{imanigooghari-etal-2023-glot500};\newline NLLB \citep{costa2022nllb}
\\[2pt]

Iranian Azerbaijani (azb)
&
---
&
---
&
Glot500 \citep{imanigooghari-etal-2023-glot500};\newline NLLB \citep{costa2022nllb}
\\[4pt]

\multicolumn{4}{l}{\cellcolor{branchbg}\textit{Kipchak Branch}} \\
\midrule

Kazakh (kaz)
&
Kaz-RoBERTa \citep{mussakhojayeva2023multilingual}
&
KAZ-LLM 8B/70B (LLaMA 3.1-based) \citep{issai2024kazllm};\newline Sherkala-Chat 8B \citep{koto2025sherkala}
&
mBERT \citep{devlin2019bert};\newline XLM-R \citep{conneau2020unsupervised};\newline Glot500 \citep{imanigooghari-etal-2023-glot500};\newline mGPT \citep{shliazhko-etal-2024-mgpt};\newline NLLB \citep{costa2022nllb}
\\[2pt]

Kyrgyz (kir)
&
KyrgyzBERT \citep{kyrgyzbert2023}
&
---
&
XLM-R \citep{conneau2020unsupervised};\newline Glot500 \citep{imanigooghari-etal-2023-glot500};\newline mGPT \citep{shliazhko-etal-2024-mgpt};\newline NLLB \citep{costa2022nllb}
\\[2pt]

Tatar (tat)
&
---
&
---
&
Glot500 \citep{imanigooghari-etal-2023-glot500};\newline mGPT \citep{shliazhko-etal-2024-mgpt};\newline NLLB \citep{costa2022nllb}
\\[2pt]

Bashkir (bak)
&
---
&
---
&
Glot500 \citep{imanigooghari-etal-2023-glot500};\newline mGPT \citep{shliazhko-etal-2024-mgpt};\newline NLLB \citep{costa2022nllb}
\\[2pt]

Crimean Tatar (crh)
&
---
&
---
&
Glot500 \citep{imanigooghari-etal-2023-glot500};\newline NLLB \citep{costa2022nllb}
\\[2pt]

Karakalpak (kaa)
&
---
&
---
&
Glot500 \citep{imanigooghari-etal-2023-glot500};\newline NLLB \citep{costa2022nllb}
\\[2pt]

Nogai (nog)
&
---
&
---
&
Glot500 \citep{imanigooghari-etal-2023-glot500};\newline NLLB \citep{costa2022nllb}
\\[2pt]

Kumyk (kum)
&
---
&
---
&
Glot500 \citep{imanigooghari-etal-2023-glot500};\newline NLLB \citep{costa2022nllb}
\\[2pt]

Karachay-Balkar (krc)
&
---
&
---
&
Glot500 \citep{imanigooghari-etal-2023-glot500};\newline NLLB \citep{costa2022nllb}
\\[4pt]

\multicolumn{4}{l}{\cellcolor{branchbg}\textit{Karluk Branch}} \\
\midrule

Uzbek (uzb)
&
UzBERT \citep{mansurov2021uzbert};\newline TahrirchiBERT \citep{tahrirchi2021bert};\newline BERTbek \citep{kuriyozov-etal-2024-bertbek}
&
---
&
mBERT \citep{devlin2019bert};\newline XLM-R \citep{conneau2020unsupervised};\newline Glot500 \citep{imanigooghari-etal-2023-glot500};\newline mGPT \citep{shliazhko-etal-2024-mgpt};\newline NLLB \citep{costa2022nllb}
\\[2pt]

Uyghur (uig)
&
---
&
---
&
XLM-R \citep{conneau2020unsupervised};\newline Glot500 \citep{imanigooghari-etal-2023-glot500};\newline NLLB \citep{costa2022nllb}
\\[4pt]

\multicolumn{4}{l}{\cellcolor{branchbg}\textit{Siberian Branch}} \\
\midrule

Sakha (sah)
&
---
&
---
&
Glot500 \citep{imanigooghari-etal-2023-glot500};\newline mGPT \citep{shliazhko-etal-2024-mgpt};\newline NLLB \citep{costa2022nllb}
\\[2pt]

Altai (alt)
&
---
&
---
&
Glot500 \citep{imanigooghari-etal-2023-glot500};\newline NLLB \citep{costa2022nllb}
\\[2pt]

Tuvan (tyv)
&
---
&
---
&
Glot500 \citep{imanigooghari-etal-2023-glot500};\newline NLLB \citep{costa2022nllb}
\\[2pt]

Khakas (kjh)
&
---
&
---
&
Glot500 \citep{imanigooghari-etal-2023-glot500};\newline NLLB \citep{costa2022nllb}
\\[4pt]

\multicolumn{4}{l}{\cellcolor{branchbg}\textit{Oghur Branch}} \\
\midrule

Chuvash (chv)
&
---
&
---
&
Glot500 \citep{imanigooghari-etal-2023-glot500};\newline mGPT \citep{shliazhko-etal-2024-mgpt};\newline NLLB \citep{costa2022nllb}
\\[4pt]

\multicolumn{4}{l}{\cellcolor{branchbg}\textit{Arghu Branch}} \\
\midrule

Khalaj (klj)
&
---
&
---
&
Glot500 \citep{imanigooghari-etal-2023-glot500};\newline NLLB \citep{costa2022nllb}
\\[4pt]

\multicolumn{4}{l}{\cellcolor{branchbg}\textit{Historical Languages}} \\
\midrule

Ottoman Turkish (ota)
&
Ottoman BERTurk \citep{schweter2020berturk}
&
---
&
Glot500 \citep{imanigooghari-etal-2023-glot500};\newline NLLB \citep{costa2022nllb}
\\[2pt]

Old Turkish (otk)
&
---
&
---
&
---
\\

\end{longtable}
\end{landscape}
\twocolumn

%% file: sections/appendix-e-challenge-sentences.tex
\section{Morphological Challenge Sentences}
\label{sec:app-challenge}

Tables~\ref{tab:cs-1}--\ref{tab:cs-10} present a parallel set of 20 morphological challenge sentences across English and 20~Turkic languages.
Each sentence exercises a specific grammatical phenomenon central to Turkic morphology.
These sentences are the diagnostic inputs of the morphological evaluation in \S\ref{subsec:eval-morph}.

\onecolumn

\newcommand{\csgroup}[2]{
  \begin{minipage}[t]{0.48\textwidth}
    \scriptsize
    \renewcommand{\arraystretch}{0.92}
    \begin{tabular}[t]{@{}l p{5.6cm}@{}}
      \multicolumn{2}{@{}l}{\normalsize\strut #1} \\
      \midrule
      #2
    \end{tabular}
  \end{minipage}%
}

\begin{table}[h!]
\centering
\csgroup{\textbf{Genitive-Possessive} \texttt{gen\_psor}}{%
eng & Ali's book is on the table. \\
alt & Алинин бичиги столдо. \\
aze & Ali'nin kitabı masadadır. \\
bak & Алиҙың китабы өҫтәлдә. \\
chv & Алин кӗнеки сӗтел ҫинче. \\
crh & Ali'niñ kitabı masada. \\
gag & Ali'nin kitabı masada. \\
kaa & Aliniń kitabı stolda. \\
kaz & Әлидің кітабы үстелде. \\
kir & Алинин китеби үстөлдө. \\
kjh & Алиның пічігі столда. \\
krc & Али́ни китабы́ столдады. \\
kum & Алини китабы столда. \\
nog & Алидинъ китабы столда. \\
sah & Али кинигэтэ остуолга. \\
tat & Алиның китабы өстәлдә. \\
tuk & Aliniň kitaby stoluň üstünde. \\
tur & Ali'nin kitabı masada. \\
tyv & Алинин ному столда. \\
uig & ئەلىنىڭ كىتابى ئۈستەلدە. \\
uzb & Alining kitobi stolda. \\
}%
\hfill
\csgroup{\textbf{Accusative Object} \texttt{acc\_object}}{%
eng & Ali saw the book. \\
alt & Али бичикти кӧрди. \\
aze & Ali kitabı gördü. \\
bak & Али китапты күрҙе. \\
chv & Али кӗнекене курнӑ. \\
crh & Ali kitanı kördi. \\
gag & Ali kitabı gördü. \\
kaa & Ali kitaptı kórdi. \\
kaz & Әли кітапты көрді. \\
kir & Али китепти көрдү. \\
kjh & Али пічікті кӧрді. \\
krc & Али́ китапны́ кёрдю́. \\
kum & Али китапны гёрдю. \\
nog & Али китапты коьрди. \\
sah & Али кинигэни көрдө. \\
tat & Али китапны күрде. \\
tuk & Ali kitaby gördi. \\
tur & Ali kitabı gördü. \\
tyv & Али номну көрген. \\
uig & ئەلى كىتابنى كۆردى. \\
uzb & Ali kitobni ko'rdi. \\
}
\caption{Challenge sentences: Genitive-Possessive and Accusative Object.}
\label{tab:cs-1}
\end{table}

\begin{table}[h!]
\centering
\csgroup{\textbf{Locative} \texttt{locative}}{%
eng & Ali is waiting at school. \\
alt & Али школдо сакып јат. \\
aze & Ali məktəbdə gözləyir. \\
bak & Али мәктәптә көтә. \\
chv & Али шкулта кӗтет. \\
crh & Ali mektepte bekley. \\
gag & Ali şkolada bekler. \\
kaa & Ali mektepte kútip otır. \\
kaz & Әли мектепте күтіп тұр. \\
kir & Али мектепте күтүп жатат. \\
kjh & Али школада полча. \\
krc & Али́ шко́лда сакъла́йды. \\
kum & Али школда гёзлеп тура. \\
nog & Али мектепте куьтеди. \\
sah & Али оскуолаҕа кэтэһэр. \\
tat & Али мәктәптә көтә. \\
tuk & Ali mekdepde garaşýar. \\
tur & Ali okulda bekliyor. \\
tyv & Али школада манап олур. \\
uig & ئەلى مەكتەپتە ساقلاۋاتىدۇ. \\
uzb & Ali maktabda kutyapti. \\
}%
\hfill
\csgroup{\textbf{Ablative} \texttt{ablative}}{%
eng & Ali left the school. \\
alt & Али школдоҥ чыкты. \\
aze & Ali məktəbdən çıxdı. \\
bak & Али мәктәптән сыҡты. \\
chv & Али шкултан тухнӑ. \\
crh & Ali mektepten çıqtı. \\
gag & Ali şkoladan çıktı. \\
kaa & Ali mektepten shıqtı. \\
kaz & Әли мектептен шықты. \\
kir & Али мектептен чыкты. \\
kjh & Али школадаң сыхты. \\
krc & Али́ шко́лдан чы́кты. \\
kum & Али школдан чыкъды. \\
nog & Али мектептен шыкты. \\
sah & Али оскуолаттан таҕыста. \\
tat & Али мәктәптән чыкты. \\
tuk & Ali mekdepden çykdy. \\
tur & Ali okuldan çıktı. \\
tyv & Али школадаан үнген. \\
uig & ئەلى مەكتەپتىن چىقتى. \\
uzb & Ali maktabdan chiqdi. \\
}
\caption{Challenge sentences: Locative and Ablative.}
\label{tab:cs-2}
\end{table}

\begin{table}[h!]
\centering
\csgroup{\textbf{Dative} \texttt{dative}}{%
eng & Ali entered the school. \\
alt & Али школго кирди. \\
aze & Ali məktəbə girdi. \\
bak & Али мәктәпкә инде. \\
chv & Али шкула кӗнӗ. \\
crh & Ali mektepke kirdi. \\
gag & Ali şkolaya girdi. \\
kaa & Ali mektepke kirdi. \\
kaz & Әли мектепке кірді. \\
kir & Али мектепке кирди. \\
kjh & Али школаа кірді. \\
krc & Али́ шко́лгъа кирди́. \\
kum & Али школгъа гирди. \\
nog & Али мектепке кирди. \\
sah & Али оскуолаҕа киирдэ. \\
tat & Али мәктәпкә керде. \\
tuk & Ali mekdebe girdi. \\
tur & Ali okula girdi. \\
tyv & Али школаже кирген. \\
uig & ئەلى مەكتەپكە كىردى. \\
uzb & Ali maktabga kirdi. \\
}%
\hfill
\csgroup{\textbf{Plural Noun} \texttt{plural\_noun}}{%
eng & The children played in the park. \\
alt & Балдар паркта ойнодылар. \\
aze & Uşaqlar parkda oynadılar. \\
bak & Балалар паркта уйнанылар. \\
chv & Ачасем паркра вылянӑ. \\
crh & Balalar parkta oynadılar. \\
gag & Uşaklar parkta oynadılar. \\
kaa & Balalar parkta oynadı. \\
kaz & Балалар саябақта ойнады. \\
kir & Балдар паркта ойношту. \\
kjh & Палалар паркта ойнадылар. \\
krc & Сабийле́ па́ркда ойна́дыла. \\
kum & Яшлар паркта ойнадылар. \\
nog & Балалар паркта ойнадылар. \\
sah & Оҕолор пааркаҕа оонньоотулар. \\
tat & Балалар паркта уйнадылар. \\
tuk & Çagalar parkda oýnadylar. \\
tur & Çocuklar parkta oynadı. \\
tyv & Уруглар паркта ойнааннар. \\
uig & بالىلار باغچىدا ئوينىدى. \\
uzb & Bolalar parkda o'ynashdi. \\
}
\caption{Challenge sentences: Dative and Plural Noun.}
\label{tab:cs-3}
\end{table}

\begin{table}[h!]
\centering
\csgroup{\textbf{1st Person Possessive} \texttt{psor\_1sg}}{%
eng & My house is big. \\
alt & Мениҥ айылым јаан. \\
aze & Mənim evim böyükdür. \\
bak & Минең өйөм ҙур. \\
chv & Манӑн кил пысӑк. \\
crh & Meniñ evim büyük. \\
gag & Benim evim büük. \\
kaa & Meniń úyim úlken. \\
kaz & Менің үйім үлкен. \\
kir & Менин үйүм чоң. \\
kjh & Минің тураам улуғ. \\
krc & Мени́ юйю́м уллу́ду. \\
kum & Мени уьюм уллу. \\
nog & Меним уьйим уьйкен. \\
sah & Мин дьиэм улахан. \\
tat & Минем өем зур. \\
tuk & Meniň öýüm uly. \\
tur & Benim evim büyük. \\
tyv & Мээң бажыңым улуг. \\
uig & مېنىڭ ئۆيۈم چوڭ. \\
uzb & Mening uyim katta. \\
}%
\hfill
\csgroup{\textbf{Past Tense} \texttt{past\_tense}}{%
eng & Ali came yesterday. \\
alt & Али тӱӱне келди. \\
aze & Ali dünən gəldi. \\
bak & Али кисә килде. \\
chv & Али ӗнер килнӗ. \\
crh & Ali tünevin keldi. \\
gag & Ali dün geldi. \\
kaa & Ali keshe keldi. \\
kaz & Әли кеше келді. \\
kir & Али кечээ келди. \\
kjh & Али тӱӱн килді. \\
krc & Али́ тюне́н келди́. \\
kum & Али тюнегюн гелди. \\
nog & Али кеше келди. \\
sah & Али бэҕэһээ кэллэ. \\
tat & Али кичә килде. \\
tuk & Ali düýn geldi. \\
tur & Ali dün geldi. \\
tyv & Али кежээ келген. \\
uig & ئەلى تۈنۈگۈن كەلدى. \\
uzb & Ali kecha keldi. \\
}
\caption{Challenge sentences: 1st Person Possessive and Past Tense.}
\label{tab:cs-4}
\end{table}

\begin{table}[h!]
\centering
\csgroup{\textbf{Present Progressive} \texttt{present\_prog}}{%
eng & Ali is coming now. \\
alt & Али эмди келип јат. \\
aze & Ali indi gəlir. \\
bak & Али хәҙер килә. \\
chv & Али халӗ килет. \\
crh & Ali şimdi kele. \\
gag & Ali şimdi geler. \\
kaa & Ali házir kelip atır. \\
kaz & Әли қазір келе жатыр. \\
kir & Али азыр келе жатат. \\
kjh & Али амды киліп турча. \\
krc & Али́ бусагъатда́ келе́ди. \\
kum & Али гьали геле. \\
nog & Али аьзир келеди. \\
sah & Али билигин кэлэр. \\
tat & Али хәзер килә. \\
tuk & Ali häzir gelýär. \\
tur & Ali şimdi geliyor. \\
tyv & Али ам кел чыр. \\
uig & ئەلى ھازىر كېلىۋاتىدۇ. \\
uzb & Ali hozir kelyapti. \\
}%
\hfill
\csgroup{\textbf{Future Tense} \texttt{future\_tense}}{%
eng & Ali will come tomorrow. \\
alt & Али эртен келер. \\
aze & Ali sabah gələcək. \\
bak & Али иртәгә киләсәк. \\
chv & Али ыран килӗ. \\
crh & Ali yarın kelecek. \\
gag & Ali yarın gelecek. \\
kaa & Ali erteń keledi. \\
kaz & Әли ертең келеді. \\
kir & Али эртең келет. \\
kjh & Али таңда килер. \\
krc & Али́ тамбла́ келли́кди. \\
kum & Али танга гележек. \\
nog & Али эртен келеек. \\
sah & Али сарсын кэлиэҕэ. \\
tat & Али иртәгә киләчәк. \\
tuk & Ali ertir geler. \\
tur & Ali yarın gelecek. \\
tyv & Али даарта кээр. \\
uig & ئەلى ئەتە كېلىدۇ. \\
uzb & Ali ertaga keladi. \\
}
\caption{Challenge sentences: Present Progressive and Future Tense.}
\label{tab:cs-5}
\end{table}

\begin{table}[h!]
\centering
\csgroup{\textbf{Negation} \texttt{negation}}{%
eng & Ali did not go to school. \\
alt & Али школго барган јок. \\
aze & Ali məktəbə getmədi. \\
bak & Али мәктәпкә барманы. \\
chv & Али шкула кайман. \\
crh & Ali mektepke barmadı. \\
gag & Ali şkolaya gitmedi. \\
kaa & Ali mektepke barmadı. \\
kaz & Әли мектепке бармады. \\
kir & Али мектепке барган жок. \\
kjh & Али школаа парбады. \\
krc & Али́ шко́лгъа барма́ды. \\
kum & Али школгъа бармады. \\
nog & Али мектепке бармады. \\
sah & Али оскуолаҕа барбата. \\
tat & Али мәктәпкә бармады. \\
tuk & Ali mekdebe gitmedi. \\
tur & Ali okula gitmedi. \\
tyv & Али школаже барбаан. \\
uig & ئەلى مەكتەپكە بارمىدى. \\
uzb & Ali maktabga bormadi. \\
}%
\hfill
\csgroup{\textbf{Question} \texttt{question}}{%
eng & Did Ali go to school? \\
alt & Али школго барды ба? \\
aze & Ali məktəbə getdi mi? \\
bak & Али мәктәпкә барҙымы? \\
chv & Али шкула кайнӑ-и? \\
crh & Ali mektepke bardı mı? \\
gag & Ali şkolaya gitti mi? \\
kaa & Ali mektepke bardı ma? \\
kaz & Әли мектепке барды ма? \\
kir & Али мектепке бардыбы? \\
kjh & Али школаа парды ба? \\
krc & Али́ шко́лгъа барды́мы? \\
kum & Али школгъа бардымы? \\
nog & Али мектепке бардыма? \\
sah & Али оскуолаҕа барда дуо? \\
tat & Али мәктәпкә бардымы? \\
tuk & Ali mekdebe gitdimi? \\
tur & Ali okula gitti mi? \\
tyv & Али школаже барды бе? \\
uig & ئەلى مەكتەپكە باردىمۇ؟ \\
uzb & Ali maktabga bordimi? \\
}
\caption{Challenge sentences: Negation and Question.}
\label{tab:cs-6}
\end{table}

\begin{table}[h!]
\centering
\csgroup{\textbf{Imperative} \texttt{imperative}}{%
eng & Ali, open the door! \\
alt & Али, эжикти ач! \\
aze & Ali, qapını aç! \\
bak & Али, ишекте ас! \\
chv & Али, алӑка уҫ! \\
crh & Ali, qapını aç! \\
gag & Ali, kapıyı aç! \\
kaa & Ali, esikti ash! \\
kaz & Әли, есікті аш! \\
kir & Али, эшикти ач! \\
kjh & Али, ізікті ас! \\
krc & Али́, эшикни́ ач! \\
kum & Али, эшикни ач! \\
nog & Али, эсикти аш! \\
sah & Али, ааны ас! \\
tat & Али, ишекне ач! \\
tuk & Ali, gapyny aç! \\
tur & Ali, kapıyı aç! \\
tyv & Али, эжикти ажыт! \\
uig & ئەلى، ئىشىكنى ئاچ! \\
uzb & Ali, eshikni och! \\
}%
\hfill
\csgroup{\textbf{Conditional} \texttt{conditional}}{%
eng & If Ali came, it would be good. \\
alt & Али келзе, јараар эди. \\
aze & Əgər Ali gəlsəydi, yaxşı olardı. \\
bak & Әгәр Али килһә, яҡшы булыр ине. \\
chv & Енчен Али килсен, аван пулнӑ пулӗччӗ. \\
crh & Eger Ali kelse, yahşı olur edi. \\
gag & Eer Ali gelseydi, ii olaceydı. \\
kaa & Eger Ali kelse, jaqsı bolar edi. \\
kaz & Егер Әли келсе, жақсы болар еді. \\
kir & Эгер Али келсе, жакшы болмок. \\
kjh & Али килзе, хоос полар. \\
krc & Али́ келсе́, иги́ боллу́къ эди. \\
kum & Али гелсе, яхшы болар эди. \\
nog & Эгер Али келсе, ийги болар эди. \\
sah & Али кэлбит буолсай, үчүгэй буолуо этэ. \\
tat & Әгәр Али килсә, яхшы булыр иде. \\
tuk & Eger Ali gelse, gowy bolardy. \\
tur & Ali gelse iyi olurdu. \\
tyv & Али келзе, эки боор ийик. \\
uig & ئەگەر ئەلى كەلسە، ياخشى بولاتتى. \\
uzb & Agar Ali kelsa, yaxshi bo'lardi. \\
}
\caption{Challenge sentences: Imperative and Conditional.}
\label{tab:cs-7}
\end{table}

\begin{table}[h!]
\centering
\csgroup{\textbf{Evidential} \texttt{evidential}}{%
eng & Ali apparently came. \\
alt & Али келип калган. \\
aze & Deyəsən, Ali gəlib. \\
bak & Күрәһең, Али килгән. \\
chv & Али килнӗ курӑнать. \\
crh & Ali kelgen kibi. \\
gag & Ali gelmiş. \\
kaa & Ali kelgen sıyaqlı. \\
kaz & Әли келіпті. \\
kir & Али келиптир. \\
kjh & Али килген полтыр. \\
krc & Али́ келгенди́. \\
kum & Али гелгендир. \\
nog & Али келген. \\
sah & Али кэлбит. \\
tat & Али килгән, ахрысы. \\
tuk & Ali gelipdir. \\
tur & Ali gelmiş. \\
tyv & Али келген-дир. \\
uig & ئەلى كېلىپتۇ. \\
uzb & Ali kelibdi. \\
}%
\hfill
\csgroup{\textbf{Passive} \texttt{passive}}{%
eng & The door was opened. \\
alt & Эжик ачылды. \\
aze & Qapı açıldı. \\
bak & Ишек асылды. \\
chv & Алӑк уҫӑлнӑ. \\
crh & Qapı açıldı. \\
gag & Kapı açıldı. \\
kaa & Esik ashıldı. \\
kaz & Есік ашылды. \\
kir & Эшик ачылды. \\
kjh & Ізік азылды. \\
krc & Эшик ачы́лды. \\
kum & Эшик ачылды. \\
nog & Эсик ашылды. \\
sah & Аан аһылынна. \\
tat & Ишек ачылды. \\
tuk & Gapy açyldy. \\
tur & Kapı açıldı. \\
tyv & Эжик ажыттынган. \\
uig & ئىشىك ئېچىلدى. \\
uzb & Eshik ochildi. \\
}
\caption{Challenge sentences: Evidential and Passive.}
\label{tab:cs-8}
\end{table}

\begin{table}[h!]
\centering
\csgroup{\textbf{Reciprocal} \texttt{reciprocal}}{%
eng & They saw each other. \\
alt & Олор бой-бойын кӧрдилӧр. \\
aze & Onlar bir-birini gördülər. \\
bak & Улар бер-береһен күрҙеләр. \\
chv & Вӗсем пӗр-пӗрне курнӑ. \\
crh & Olar bir-birini kördiler. \\
gag & Onnar biri-birini gördülär. \\
kaa & Olar bir-birin kórdi. \\
kaz & Олар бір-бірін көрді. \\
kir & Алар бири-бирин көрүштү. \\
kjh & Олар позы-позын кӧрістілер. \\
krc & Ала́ бир-бирлерин кёрдюле́. \\
kum & Олар бир-бирин гёрдюлер. \\
nog & Олар бир-бирин коьрдилер. \\
sah & Кинилэр бэйэ-бэйэлэрин көрүстүлэр. \\
tat & Алар бер-берсен күрделәр. \\
tuk & Olar biri-birini gördüler. \\
tur & Onlar birbirini gördü. \\
tyv & Олар бот-боттарын көргеннер. \\
uig & ئۇلار بىر-بىرىنى كۆردى. \\
uzb & Ular bir-birlarini ko'rdilar. \\
}%
\hfill
\csgroup{\textbf{Pronouns} \texttt{pronouns}}{%
eng & We know you. \\
alt & Бис сени билерис. \\
aze & Biz səni tanıyırıq. \\
bak & Беҙ һине беләбеҙ. \\
chv & Эпир сана пӗлетпӗр. \\
crh & Biz seni tanıyırmız. \\
gag & Biz seni tanıyırız. \\
kaa & Biz seni tanıymız. \\
kaz & Біз сені танимыз. \\
kir & Биз сени тааныйбыз. \\
kjh & Піс сені пілчебіс. \\
krc & Биз сени́ таныйбыз. \\
kum & Биз сени таныйбыз. \\
nog & Биз сени таниймыз. \\
sah & Биһиги эйигин билэбит. \\
tat & Без сине беләбез. \\
tuk & Biz seni tanaýarys. \\
tur & Biz seni tanıyoruz. \\
tyv & Бис сени таныыр бис. \\
uig & بىز سېنى تونۇيمىز. \\
uzb & Biz seni taniymiz. \\
}
\caption{Challenge sentences: Reciprocal and Pronouns.}
\label{tab:cs-9}
\end{table}

\begin{table}[h!]
\centering
\csgroup{\textbf{Numeral} \texttt{numeral}}{%
eng & Three students came. \\
alt & Ӱч студент келди. \\
aze & Üç tələbə gəldi. \\
bak & Өс студент килде. \\
chv & Виҫӗ студент килнӗ. \\
crh & Üç student keldi. \\
gag & Üç öğrenci geldi. \\
kaa & Úsh student keldi. \\
kaz & Үш студент келді. \\
kir & Үч студент келди. \\
kjh & Ӱс студент килді. \\
krc & Юч студент келди́. \\
kum & Уьч студент гелди. \\
nog & Уьш студент келди. \\
sah & Үс үөрэнээччи кэллэ. \\
tat & Өч студент килде. \\
tuk & Üç talyp geldi. \\
tur & Üç öğrenci geldi. \\
tyv & Үш сургуул келген. \\
uig & ئۈچ ئوقۇغۇچى كەلدى. \\
uzb & Uchta talaba keldi. \\
}%
\hfill
\csgroup{\textbf{Converb} \texttt{converb}}{%
eng & Ali spoke while laughing. \\
alt & Али кӱлӱп куучындады. \\
aze & Ali gülərək danışdı. \\
bak & Али көлөп һөйләне. \\
chv & Али кулса калаҫнӑ. \\
crh & Ali külerek laf etti. \\
gag & Ali gülerek laf etti. \\
kaa & Ali kúlip sóyledi. \\
kaz & Әли күліп сөйледі. \\
kir & Али күлүп сүйлөдү. \\
kjh & Али кӱліп чоохтады. \\
krc & Али́ кюле́ туруп сёлешди́. \\
kum & Али кюлеп сёйледи. \\
nog & Али куьлип соьйледи. \\
sah & Али күлэн саҥарда. \\
tat & Али көлеп сөйләде. \\
tuk & Ali ýylgyryp gürledi. \\
tur & Ali gülerek konuştu. \\
tyv & Али хүлүмзүрүп чугаалаан. \\
uig & ئەلى كۈلۈپ سۆزلىدى. \\
uzb & Ali kulib gapirdi. \\
}
\caption{Challenge sentences: Numeral and Converb.}
\label{tab:cs-10}
\end{table}

\twocolumn

%% file: sections/appendix-f-script-results.tex
\section{Script-Layer Results}
\label{sec:app-script-results}

Tables~\ref{tab:translit-eval}--\ref{tab:cts-fertility} give the detailed results summarized in \S\ref{subsec:eval-scripts} and \S\ref{subsec:eval-cts}.

\begin{table}[h]
  \centering
  \footnotesize
  \setlength{\tabcolsep}{2.6pt}
  \begin{tabular}{@{}llrrrr@{}}
    \toprule
    \textbf{Lang.} & \textbf{Round trip} & \textbf{Words} & \textbf{Word\%} & \textbf{CER} & \textbf{Purity} \\
    \midrule
    aze & Latn$\to$Cyrl$\to$Latn & 784 & 100.0 & 0.00 & 100.0 \\
    tuk & Latn$\to$Cyrl$\to$Latn &  62 & 100.0 & 0.00 & 100.0 \\
    uzb & Latn$\to$Cyrl$\to$Latn & 913 & 100.0 & 0.00 & 100.0 \\
    crh & Latn$\to$Cyrl$\to$Latn & 822 &  97.6 & 0.37 & 100.0 \\
    kaa & Latn$\to$Cyrl$\to$Latn & 272 & 100.0 & 0.00 &  99.7 \\
    kaz & Cyrl$\to$Latn$\to$Cyrl &  65 & 100.0 & 0.00 & 100.0 \\
    tat & Cyrl$\to$Latn$\to$Cyrl &  87 & 100.0 & 0.00 & 100.0 \\
    uig & Arab$\to$Latn$\to$Arab &  61 & 100.0 & 0.00 & 100.0 \\
    uig & Arab$\to$Cyrl$\to$Arab &  61 & 100.0 & 0.00 & 100.0 \\
    uig & Arab$\to$CTS$\to$Arab  &  61 & 100.0 & 0.00 & 100.0 \\
    \bottomrule
  \end{tabular}
  \caption{Transliteration round-trip consistency. \textbf{Word\%}: words recovered exactly after the round trip; \textbf{CER}: character error rate (\%) of the round trip; \textbf{Purity}: share (\%) of letters in the intermediate output that belong to the target script.}
  \label{tab:translit-eval}
\end{table}

\begin{table}[h]
  \centering
  \footnotesize
  \setlength{\tabcolsep}{4pt}
  \begin{tabular}{@{}lrrrr@{}}
    \toprule
    & \multicolumn{2}{c}{\textbf{Challenge}} & \multicolumn{2}{c}{\textbf{TueCL}} \\
    \cmidrule(lr){2-3} \cmidrule(lr){4-5}
    \textbf{Lang.} & \textbf{Full} & \textbf{Turkic} & \textbf{Full} & \textbf{Turkic} \\
    \midrule
    tur & 100 & 100 & 100.0 & 100.0 \\
    aze &  90 &  90 &  89.2 &  89.2 \\
    tuk &  80 &  80 & -- & -- \\
    gag &  10 &  10 & -- & -- \\
    kaz & 100 & 100 & -- & -- \\
    kir & 100 & 100 &  96.5 &  97.7 \\
    tat &  90 &  95 & 100.0 & 100.0 \\
    bak & 100 & 100 & -- & -- \\
    crh &  45 &  45 &  51.7 &  52.4 \\
    kaa &   0 &   0 &  34.3 &  34.3 \\
    nog &  30 &  35 & -- & -- \\
    kum &  25 &  25 &  84.2 &  89.5 \\
    krc &  80 &  80 & -- & -- \\
    uzb & 100 & 100 &  97.3 &  98.0 \\
    uig & 100 & 100 & -- & -- \\
    sah &  95 &  95 & 100.0 & 100.0 \\
    alt &  40 &  55 & -- & -- \\
    tyv &  45 &  45 & -- & -- \\
    kjh &  25 &  35 & -- & -- \\
    chv &  90 &  95 & -- & -- \\
    \midrule
    Micro avg. & 67.2 & 69.2 & 85.1 & 85.7 \\
    Macro avg. & 67.2 & 69.2 & 83.7 & 84.6 \\
    \bottomrule
  \end{tabular}
  \caption{Language identification accuracy (\%) of GlotLID with its full label space (2,102 labels) and restricted to TurkicNLP's Turkic labels (30 labels), on the 20 challenge sentences per language and the natural TueCL sentences (826 sentences in 9 languages; Sakha 3, Tatar 5, Kumyk 19, Karakalpak 35 sentences).}
  \label{tab:langid-eval}
\end{table}

\begin{table}[h]
  \centering
  \scriptsize
  \setlength{\tabcolsep}{2.1pt}
  \begin{tabular}{@{}lrrrrrrrr@{}}
    \toprule
    & \multicolumn{2}{c}{\textbf{Qwen3.5}} & \multicolumn{2}{c}{\textbf{LLaMA 3}} & \multicolumn{2}{c}{\textbf{OLMo 3}} & \multicolumn{2}{c}{\textbf{GPT-5.2}} \\
    \cmidrule(lr){2-3} \cmidrule(lr){4-5} \cmidrule(lr){6-7} \cmidrule(lr){8-9}
    \textbf{Lang.} & \textbf{Nat.} & \textbf{CTS} & \textbf{Nat.} & \textbf{CTS} & \textbf{Nat.} & \textbf{CTS} & \textbf{Nat.} & \textbf{CTS} \\
    \midrule
    kaz & 2.86 & 2.28 & 3.51 & 2.32 & 4.05 & 2.55 & 2.37 & 2.29 \\
    kir & 2.84 & 2.16 & 3.48 & 2.31 & 3.80 & 2.59 & 2.52 & 2.12 \\
    tat & 3.13 & 2.37 & 4.13 & 2.59 & 4.46 & 2.71 & 2.51 & 2.40 \\
    bak & 3.35 & 2.33 & 4.68 & 2.57 & 5.00 & 2.68 & 2.54 & 2.35 \\
    nog & 2.92 & 2.02 & 3.08 & 2.08 & 3.48 & 2.27 & 2.65 & 1.98 \\
    kum & 2.76 & 2.10 & 3.06 & 2.23 & 3.61 & 2.42 & 2.56 & 2.08 \\
    krc & 2.25 & 1.85 & 2.52 & 1.88 & 2.94 & 2.01 & 2.23 & 1.80 \\
    uig & 4.51 & 2.26 & 5.62 & 2.38 & 6.85 & 2.64 & 3.38 & 2.26 \\
    sah & 3.65 & 2.35 & 4.21 & 2.43 & 4.78 & 2.63 & 3.48 & 2.37 \\
    alt & 2.83 & 2.06 & 3.08 & 2.18 & 3.63 & 2.38 & 2.63 & 1.97 \\
    tyv & 2.67 & 2.11 & 3.09 & 2.18 & 3.59 & 2.23 & 2.44 & 2.02 \\
    kjh & 2.79 & 2.13 & 3.16 & 2.16 & 3.94 & 2.37 & 2.67 & 2.10 \\
    chv & 3.52 & 2.03 & 3.66 & 2.18 & 4.09 & 2.29 & 3.42 & 2.03 \\
    \midrule
    Avg.\ non-Latin (13) & 3.08 & \textbf{2.16} & 3.64 & \textbf{2.27} & 4.17 & \textbf{2.45} & 2.72 & \textbf{2.14} \\
    Avg.\ Latin (7)      & 2.13 & 2.15 & 2.28 & 2.26 & 2.47 & 2.48 & 2.10 & 2.14 \\
    Turkish              & 1.95 & 1.95 & 2.03 & 2.03 & 2.33 & 2.33 & 1.95 & 1.95 \\
    \bottomrule
  \end{tabular}
  \caption{Tokenizer fertility (tokens per word) on the 20 parallel challenge sentences in the native script (Nat.) and after conversion to the Common Turkic Alphabet (CTS) by TurkicNLP, for the 13 languages written in Cyrillic or Perso-Arabic script. The Latin-script group covers Turkish, Azerbaijani, Turkmen, Gagauz, Crimean Tatar, Karakalpak, and Uzbek.}
  \label{tab:cts-fertility}
\end{table}

%% file: sections/appendix-g-morphseg.tex
\section{Morpheme Segmentation Evaluation}
\label{sec:app-morphseg}

\paragraph{Test set.} We built a breadth test set of 483 inflected and derived word forms for the 20 languages that have both an Apertium transducer and the neural analyzer (Ottoman Turkish, Khalaj, and South Azerbaijani lack one of the two). Words are grouped into 13 categories that cover the main slots of the Turkic suffix chain: plural; case (dative, locative, ablative, accusative, genitive, instrumental); possessive; plural+possessive+case chains; past tense+person; present/aorist+person; future; negation; participles; converbs; voice and derivation (causative, passive, \textit{-lIk}, \textit{-CI}, \textit{-sIz}, \textit{-lI}); unsegmentable words (to measure over-segmentation); and, for Turkish, proper nouns with apostrophe suffixes. The items are built from one shared core vocabulary (\textit{book, house, child, school, city, friend, fish, salt}; \textit{go, come, write, read, see}) rendered in each language's standard orthography and primary script, so that results are comparable across languages. Each language has 14--34 items (1--6 per category); categories absent from a language's morphology (e.g., the Kyrgyz and Uzbek synthetic future) or for which we could not establish a reliable form are omitted. Table~\ref{tab:morphseg-examples} shows one item per category for five languages.

\paragraph{Annotation.} Every item has a gold segmentation and one label per morpheme (\textsc{stem}, \textsc{pl}, \textsc{dat}, \textsc{poss.1sg}, \textsc{pst}, \textsc{neg}, \textsc{ptcp}, \textsc{caus}, \ldots). Buffer consonants and linking vowels belong to the following suffix (\textit{oku·yacak}, \textit{kapı·sı}), stem-final alternations stay in the stem (\textit{kitab·ı}), and zero morphemes are not segmented (\textit{gel·di}, 3\textsc{sg} unmarked). A gold label also matches a more specific predicted label (\textsc{pst} ${\sim}$ \textsc{pst.3sg}, \textsc{ptcp} ${\sim}$ \textsc{ptcp.pst}). Gold annotations follow reference grammars; we consider Turkish, Azerbaijani, Kazakh, Kyrgyz, Tatar, and Uzbek reliable (\textit{high}), Turkmen, Bashkir, Crimean Tatar, Karakalpak, and Uyghur \textit{medium}, and the remaining nine languages \textit{low} confidence; the latter are drafts awaiting native-speaker verification.

\paragraph{Configurations and metrics.} We run the segmenter with the neural analyzer only, with the Apertium transducer only, and in its default hybrid mode, in which every transducer reading is converted into a segmentation, the reading that explains the whole word (ties broken by agreement with the neural lemma and part of speech) is kept, and the neural analysis is used only when the transducer has no better reading. We report the share of words segmented exactly as in the gold standard (\textbf{Exact}), the F1 of internal morpheme boundaries (\textbf{BF1}), and the share of gold morphemes found with the correct span and label (\textbf{Lab}). The test set was also used while developing the segmenter, so absolute numbers are optimistic.

\begin{table}[h]
  \centering
  \footnotesize
  \setlength{\tabcolsep}{2.2pt}
  \begin{tabular}{@{}l@{\hspace{3pt}}c r rr rr rrr@{}}
    \toprule
    & & & \multicolumn{2}{c}{\textbf{Neural}} & \multicolumn{2}{c}{\textbf{Apertium}} & \multicolumn{3}{c}{\textbf{Hybrid}} \\
    \cmidrule(lr){4-5} \cmidrule(lr){6-7} \cmidrule(l){8-10}
    \textbf{Lang.} & \textbf{C} & $n$ & \textbf{Ex.} & \textbf{BF1} & \textbf{Ex.} & \textbf{BF1} & \textbf{Ex.} & \textbf{BF1} & \textbf{Lab} \\
    \midrule
    tur & H & 34 & 74 & 87 & 94 & 98 & \textbf{97} & 99 & 96 \\
    aze & H & 30 & 67 & 76 & 23 & 21 & \textbf{70} & 80 & 65 \\
    tuk & M & 30 & 67 & 80 & 87 & 95 & \textbf{93} & 97 & 88 \\
    gag & L & 16 & 88 & 92 & 81 & 89 & \textbf{94} & 97 & 92 \\
    kaz & H & 34 & 82 & 91 & 88 & 96 & \textbf{97} & 99 & 98 \\
    kir & H & 30 & 67 & 76 & 80 & 89 & \textbf{87} & 94 & 87 \\
    tat & H & 32 & 66 & 82 & 94 & 95 & \textbf{97} & 99 & 96 \\
    bak & M & 31 & 48 & 71 & 77 & 87 & \textbf{84} & 94 & 86 \\
    crh & M & 32 & 81 & 92 & \textbf{100} & 100 & \textbf{100} & 100 & 97 \\
    kaa & M & 30 & 73 & 86 & 90 & 94 & \textbf{100} & 100 & 96 \\
    nog & L & 18 & 56 & 75 & 67 & 79 & \textbf{89} & 94 & 90 \\
    kum & L & 16 & 69 & 83 & \textbf{88} & 94 & 81 & 91 & 80 \\
    krc & L & 14 & 64 & 79 & \textbf{93} & 90 & 86 & 87 & 79 \\
    uzb & H & 32 & 78 & 90 & 88 & 88 & \textbf{94} & 95 & 91 \\
    uig & M & 18 & 61 & 65 & 83 & 92 & \textbf{94} & 97 & 95 \\
    sah & L & 20 & 50 & 68 & \textbf{80} & 90 & 70 & 84 & 72 \\
    alt & L & 16 & 75 & 79 & 38 & 56 & \textbf{88} & 91 & 82 \\
    tyv & L & 16 & 62 & 73 & \textbf{75} & 82 & 62 & 73 & 65 \\
    kjh & L & 17 & \textbf{71} & 83 & 29 & 27 & \textbf{71} & 83 & 67 \\
    chv & L & 17 & 41 & 34 & \textbf{88} & 92 & \textbf{88} & 88 & 84 \\
    \midrule
    \textbf{All} & & 483 & 68 & 81 & 79 & 87 & \textbf{89} & 94 & 87 \\
    \bottomrule
  \end{tabular}
  \caption{Morpheme segmentation per language (\%). \textbf{C}: confidence of the gold annotation (High, Medium, Low). \textbf{Ex.}: exact segmentation; \textbf{BF1}: morpheme-boundary F1; \textbf{Lab}: morphemes with correct span and label. \textbf{All}: over all 483 words. Bold: best exact score per language.}
  \label{tab:morphseg-lang}
\end{table}

\begin{table}[h]
  \centering
  \footnotesize
  \setlength{\tabcolsep}{2.5pt}
  \begin{tabular}{@{}lrrrr@{}}
    \toprule
    \textbf{Category} & $n$ & \textbf{Neural} & \textbf{Apertium} & \textbf{Hybrid} \\
    \midrule
    Plural                    &  47 & 94 & 79 & \textbf{96} \\
    Case                      &  92 & 91 & 83 & \textbf{97} \\
    Possessive                &  47 & 77 & 79 & \textbf{87} \\
    Pl.+poss.+case            &  25 & 84 & 80 & \textbf{92} \\
    Past + person             &  40 & 60 & 80 & \textbf{88} \\
    Present/aorist + person   &  24 & 50 & 63 & \textbf{83} \\
    Future                    &  15 & 60 & 67 & \textbf{87} \\
    Negation                  &  30 & 47 & 80 & \textbf{87} \\
    Participle                &  30 & 57 & 73 & \textbf{77} \\
    Converb                   &  32 & 50 & \textbf{84} & \textbf{84} \\
    Voice and derivation      &  49 & 12 & 63 & \textbf{76} \\
    Unsegmentable             &  50 & 86 & \textbf{96} & 94 \\
    Proper noun + apostrophe  &   2 & \textbf{100} & \textbf{100} & \textbf{100} \\
    \midrule
    \textbf{All}              & 483 & 68 & 79 & \textbf{89} \\
    \bottomrule
  \end{tabular}
  \caption{Exact segmentation (\%) per category over all 20 languages.}
  \label{tab:morphseg-cat}
\end{table}

\paragraph{Results.} The two analyzers are complementary (Table~\ref{tab:morphseg-lang}). The neural analyzer covers every word but often misplaces the stem boundary, especially in verbs, and almost never splits derivational suffixes (12\%). The transducers segment verbs and derivations well but leave words outside their lexicons unanalysed, which is fatal where coverage is small (Azerbaijani, Khakas, Altai). Combining them improves exact segmentation from 68\% (neural) and 79\% (Apertium) to 89\%; the hybrid is best or tied in every category except unsegmentable words, where the transducer alone over-segments slightly less (96\% vs.\ 94\%; Table~\ref{tab:morphseg-cat}). The remaining errors concentrate in the low-coverage languages, where the hybrid falls back to the neural analysis, and in derivation and participles. Gold data, predictions for every word, and the evaluation script are part of the released code.

\begin{table*}[t]
  \centering
  \footnotesize
  \setlength{\tabcolsep}{3pt}
  \begin{tabular}{@{}lllll@{}}
    \toprule
    \textbf{Category} & \textbf{Turkish} & \textbf{Kazakh} & \textbf{Uzbek} & \textbf{Sakha} \\
    \midrule
    Plural            & kitap·lar & кітап·тар & kitob·lar & кинигэ·лэр \\
    Case              & ev·de & мектеп·ке & maktab·ga & оскуола·ҕа \\
    Possessive        & kapı·sı & қала·сы & bola·si & оҕо·то \\
    Pl.+poss.+case    & kitap·lar·ımız·dan & кітап·тар·ымыз·дан & kitob·lar·imiz·dan & кинигэ·лэр·бит \\
    Past + person     & gel·di·m & бар·ды·м & bor·di·m & бар·ды·м \\
    Present + person  & gid·iyor·um & бар·а·мын & bor·a·man & бар·а·бын \\
    Future            & oku·yacak·lar & бар·ар·мын & -- & бар·ыа·м \\
    Negation          & yaz·ma·dı·m & бар·ма·ды·м & bor·ma·di·m & -- \\
    Participle        & oku·yan & жаз·ған & yoz·gan & бар·быт \\
    Converb           & gel·ip & кел·іп & kel·ib & бар·ан \\
    Voice/derivation  & yaz·dır·dı, balık·çı & жаз·ыл·ды, тұз·сыз & yoz·dir·di, baliq·chi & оҕо·лоох \\
    Unsegmentable     & güzel & жақсы & yaxshi & үчүгэй \\
    \bottomrule
  \end{tabular}
  \caption{Example items of the morpheme segmentation test set (gold segmentation, morphemes separated by ·). The full set of 483 items in 20 languages is released with the code.}
  \label{tab:morphseg-examples}
\end{table*}